\documentclass[pdflatex,sn-mathphys-num]{sn-jnl}% Math and Physical Sciences Numbered Reference Style 
\usepackage{graphicx}%
\usepackage{multirow}%
\usepackage{amsmath,amssymb,amsfonts}%
\usepackage{amsthm}%
\usepackage{mathrsfs}%
\usepackage[title]{appendix}%
\usepackage{xcolor}%
\usepackage{textcomp}%
\usepackage{manyfoot}%
\usepackage{booktabs}%
\usepackage{algorithm}%
\usepackage{algorithmicx}%
\usepackage{algpseudocode}%
\usepackage{listings}%
\usepackage{bm}
\newcommand{\StimSystem}{S1}
\newcommand{\SpontPattern}{S2}
\newcommand{\StimPattern}{S3}
\newcommand{\Direction}{S7}
\newcommand{\SupVidESNType}{S8}
\newcommand{\SupMatSensorTesting}{S9}

\newcommand{\SupMatConfusion}{S11}
\newcommand{\SupMatXiaoTesting}{S12}
\newcommand{\SupVidViableRC}{S13}
\newcommand{\DLC}{S14}
\newcommand{\Pulse}{S15}
\newcommand{\ESPindex}{S4}
\newcommand{\StaticsP}{S5}
\newcommand{\StaticsJ}{S6}

\theoremstyle{thmstyleone}%
\theoremstyle{thmstyletwo}%

\theoremstyle{thmstylethree}%

\begin{document}

\title[A Jellyfish Cyborg: Exploiting Natural Embodied Intelligence as Soft Robots]{A Jellyfish Cyborg: Exploiting Natural Embodied Intelligence as Soft Robots}


\author*[1]{\fnm{Dai} \sur{Owaki}}\email{owaki@tohoku.ac.jp}

\author[2]{\fnm{Max} \sur{Austin}}\email{austin@isi.imi.i.u-tokyo.ac.jp}

\author[3]{\fnm{Shuhei} \sur{Ikeda}}\email{s.ikeda@kamo-kurage.jp}

\author[3]{\fnm{Kazuya} \sur{Okuizumi}}\email{kazu@kamo-kurage.jp}

\author[2]{\fnm{Kohei} \sur{Nakajima}}\email{k-nakajima@isi.imi.i.u-tokyo.ac.jp}

\affil*[1]{\orgdiv{Department of Robotics}, \orgname{Graduate School of Engineering,Tohoku University}, \orgaddress{\street{6-6-01 Aoba, Aramaki, Aoba-ku}, \city{Sendai}, \postcode{980-8579}, \country{Japan}}}

\affil[2]{\orgdiv{Graduate School of Information Science and Technology}, \orgname{The University of Tokyo}, \orgaddress{\street{7-3-1 Hongo, Bunkyo-ku}, \city{Tokyo}, \postcode{113-8656}, \country{Japan}}}

\affil[3]{\orgname{Kamo Aquarium}, \orgaddress{\street{657-1 Okubo, Imaizumi}, \city{Tsuruoka}, \postcode{997-1206}, \state{Yamagata}, \country{Japan}}}


%ver.2024.10.21.
\abstract{
Jellyfish cyborgs present a promising avenue for soft robotic systems, leveraging the natural energy-efficiency and adaptability of biological systems. Here we demonstrate a novel approach to predicting and controlling jellyfish locomotion by harnessing the natural embodied intelligence of these animals. We developed an integrated muscle electrostimulation and 3D motion capture system to quantify both spontaneous and stimulus-induced behaviors in {\it Aurelia coerulea} jellyfish. Using Reservoir Computing, a machine learning framework, we successfully predicted future movements based on the current body shape and natural dynamic patterns of the jellyfish. Our key findings include the first investigation of self-organized criticality in jellyfish swimming motions and the identification of optimal stimulus periods (1.5 and 2.0 seconds) for eliciting coherent and predictable swimming behaviors. These results suggest that the jellyfish body motion, combined with targeted electrostimulation, can serve as a computational resource for predictive control. Our findings pave the way for developing jellyfish cyborgs capable of autonomous navigation and environmental exploration, with potential applications in ocean monitoring and pollution management.
}

\maketitle

\section*{}\label{intro}

At a fundamental level, the purpose of a cyborg—defined as the fusion of a living biological system with mechatronic enhancements—is to leverage the natural intelligence that arises from the interaction between an animal and its natural environment for synthetic applications. Animals have been optimized through a long evolutionary process to interact with their native ecosystems, allowing them to fit into specific ecological niches. Numerous studies have demonstrated that behaviors such as locomotion \cite{whelan1996control,beal2006passive,Jayaram2018Transition}, learning \cite{Wells1959Touch,carducci2018tactile}, and manipulation \cite{carducci2018tactile,demery2011vision} are, at least in part, ingrained in the physiological structures of animals. Therefore, the effective design and control of the mechatronic components of a cyborg should integrate seamlessly with the complex systems of muscles, nerves, and sensory receptors, prioritizing the preservation of the natural ``embodied intelligence."

Several cyborgs \cite{Suzumori2023} have been developed with the capability to maneuver rapidly and robustly through complex terrains while simultaneously performing highly technical cognitive tasks that would be challenging to train an animal to accomplish independently \cite{wu2016cyborg,menciassi2020biohybrid,romano2019review}. Current research has shown that these systems can navigate across various locomotion domains, including terrestrial, aquatic, and aerial environments \cite{romano2019review}. They have also demonstrated potential in navigational tasks, with studies highlighting the ability to steer a robot using an animal's sensory organs \cite{ando2020insect} and to control an animal's movements with an electronic controller \cite{Sato2009RemoteRC,VoDoan2017AnUA,nguyen2023efficient,siljak2022cyborg,latif2012line}. While such cyborgs illustrate the broad applications of bio-hybrid machines, they have yet to achieve the same level of sophistication as their biological counterparts or match the peak performance of state-of-the-art bio-inspired robots \cite{Pfeifer2007SelfOrganizationEA,Ijspeert2014,vasquez2023design,ilami2021materials}.

One group of animals that has become a source of inspiration for soft swimming robots is jellyfish. Jellyfish belong to the subphylum Medusozoa and can be found in various shapes and sizes, all sharing the same general swimming mechanism: pulsatile jetting via an umbrellar structure \cite{megill2002biomechanics,costello2021hydrodynamics}. Despite their apparent simplicity, jellyfish are highly efficient swimmers \cite{miles2019don,neil2018jet}, capable of intentional turning maneuvers \cite{costello2021hydrodynamics} and equipped with sophisticated self-healing mechanisms \cite{sinigaglia2020pattern}. These characteristics have inspired the development of several bio-inspired robots \cite{nir2012jellyfish,yeom2009biomimetic,wang2023versatile,ko2012jellyfish,villanueva2011biomimetic}, including a few jellyfish cyborgs. Research on these cyborgs has shown that by applying electrical stimuli to the jellyfish's muscles, researchers can successfully elicit jetting behavior and subsequently increase the swimming speed of the jellyfish \cite{xu2020low}. Furthermore, these cyborgs were made controllable by attaching a small control electronics to the subumbrellar portion of the bell, allowing for feedforward speed modulation \cite{xu2020field}. While these systems represent impressive initial examples of jellyfish cyborgs, they have yet to demonstrate steering behaviors or the ability to predict the animal's body motions—both of which are necessary for more complex navigation tasks.

The prediction and maneuver of jellyfish cyborgs present pivotal challenges. Swimming behaviors in jellyfish arise from complex fluid-structure interactions in a soft body immersed in water. This modeling is further complicated by the need to predict the interactions between the jellyfish's spontaneous neural responses and electrical stimuli. Additionally, as is often the case with cyborgs, the small size of jellyfish limits computational resources. In particular, any added mass required for computing systems may exceed the animal's small force production— a factor that significantly contributes to the jellyfish's efficiency—while also necessitating the requisite softness of the body.

One possible method to address these issues is to utilize the jellyfish's body as a computational resource through Physical Reservoir Computing (PRC) \cite{Nakajima2020,Nakajima2021}. This approach leverages the diverse state trajectories of a natural nonlinear dynamical system, referred to as a \textit{reservoir}, to solve temporal machine learning tasks that typically require memory. For a system to function effectively as a reservoir, it must meet certain minimal conditions, including the ability to reproduce responses against identical input streams; this characteristic is known as the Echo State Property (ESP) \cite{jaeger2001echo}. Previous studies have reported the use of biological systems for PRC, including cultured neural networks \cite{Kubota2019}, plants \cite{Pieters2022}, the unicellular organism \textit{Tetrahymena thermophila} \cite{Ushio2023}, and brain organoids \cite{cai2023brain}. By applying the PRC methodology to a biological dynamical system, it may be possible to simultaneously process sensory information, compute planning algorithms, and determine subsequent control inputs by utilizing the animal's natural and spontaneous motion alongside a small, simple circuit.

In this study, we aim to develop a pathway for designing and controlling jellyfish cyborgs by leveraging the animal's embodied intelligence. To achieve this, it is crucial to understand how this intelligence manifests and how our interventions can synchronize with it. We here build a system that enables the exploration of the dynamics of the jellyfish and their responses to external muscle electrostimulation. We then analyze the predictive and computational abilities of the animal and develop a computationally inexpensive predictive model of the jellyfish. Finally, we discuss the implications of these results and how they may be utilized to create jellyfish cyborgs capable of switching between exploratory natural behaviors and highly controllable pulsatile motions.

\section*{Results}\label{results}

\begin{figure}[t]
    \centering
    \includegraphics[width=.99\linewidth]{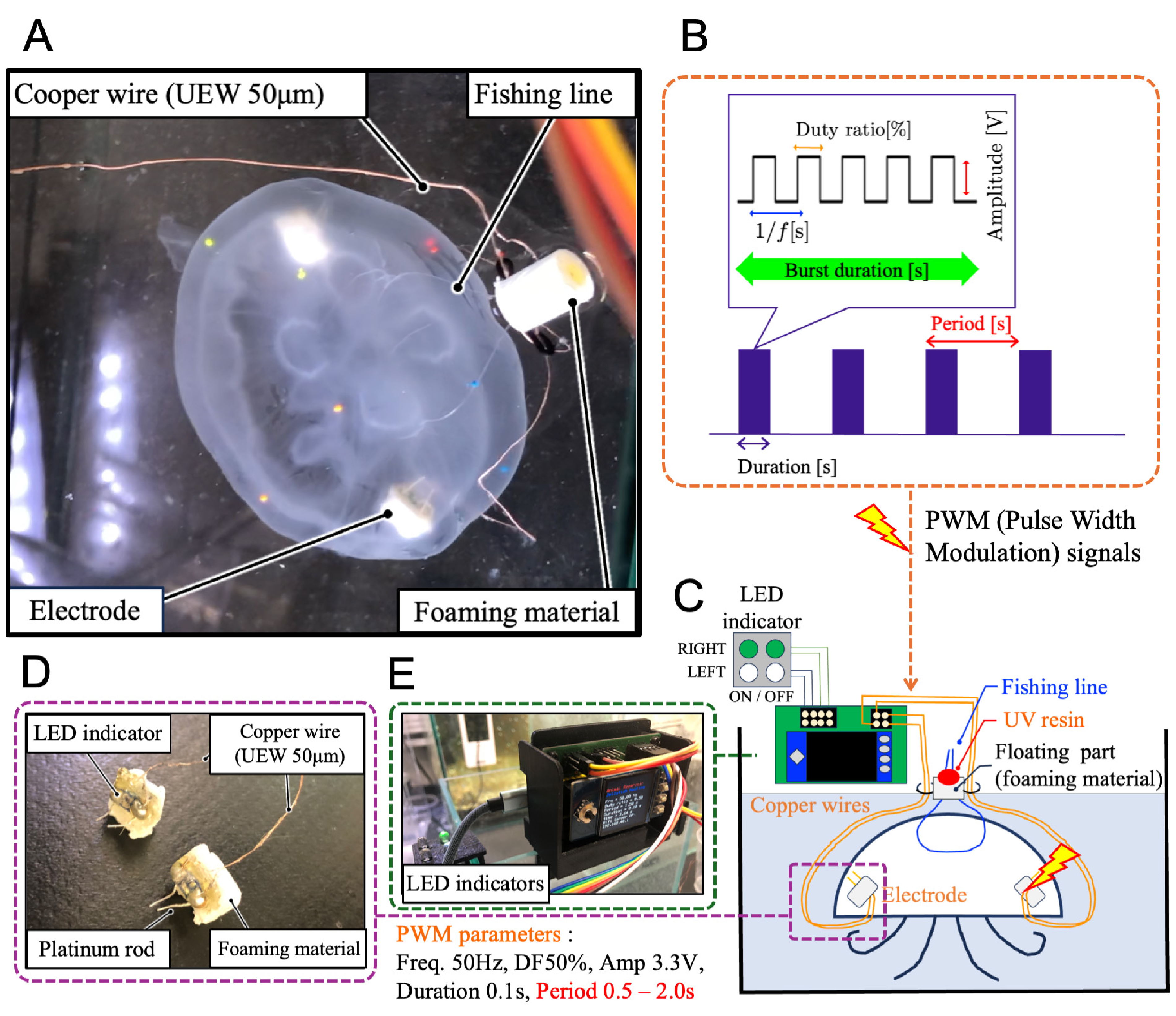}
    \caption{Overview of Jellyfish Cyborg System: ({\bf A}) Electrode-mounted jellyfish with a floating tethered system specially designed for this study. ({\bf B}) Pulse width modulation (PWM) signals used for stimulating the jellyfish muscles, which replicate the neural commands. ({\bf C}) Schematic diagram of the proposed floating tethered system. ({\bf D}) Electrodes equipped with LED stimulus indicators and wires. ({\bf E}) Custom-built electrical stimulator, utilizing a Raspberry Pi Pico W. The parameters of the PWM signals are adjusted by a Python program that integrated into the stimulator.}
    \label{fig:system}
  %\vspace{-7mm}
\end{figure}
\subsection*{Jellyfish Cyborg System for Embedding Embodied Intelligence}
\begin{figure}[t]
      \centering
      \includegraphics[width=.99\linewidth]{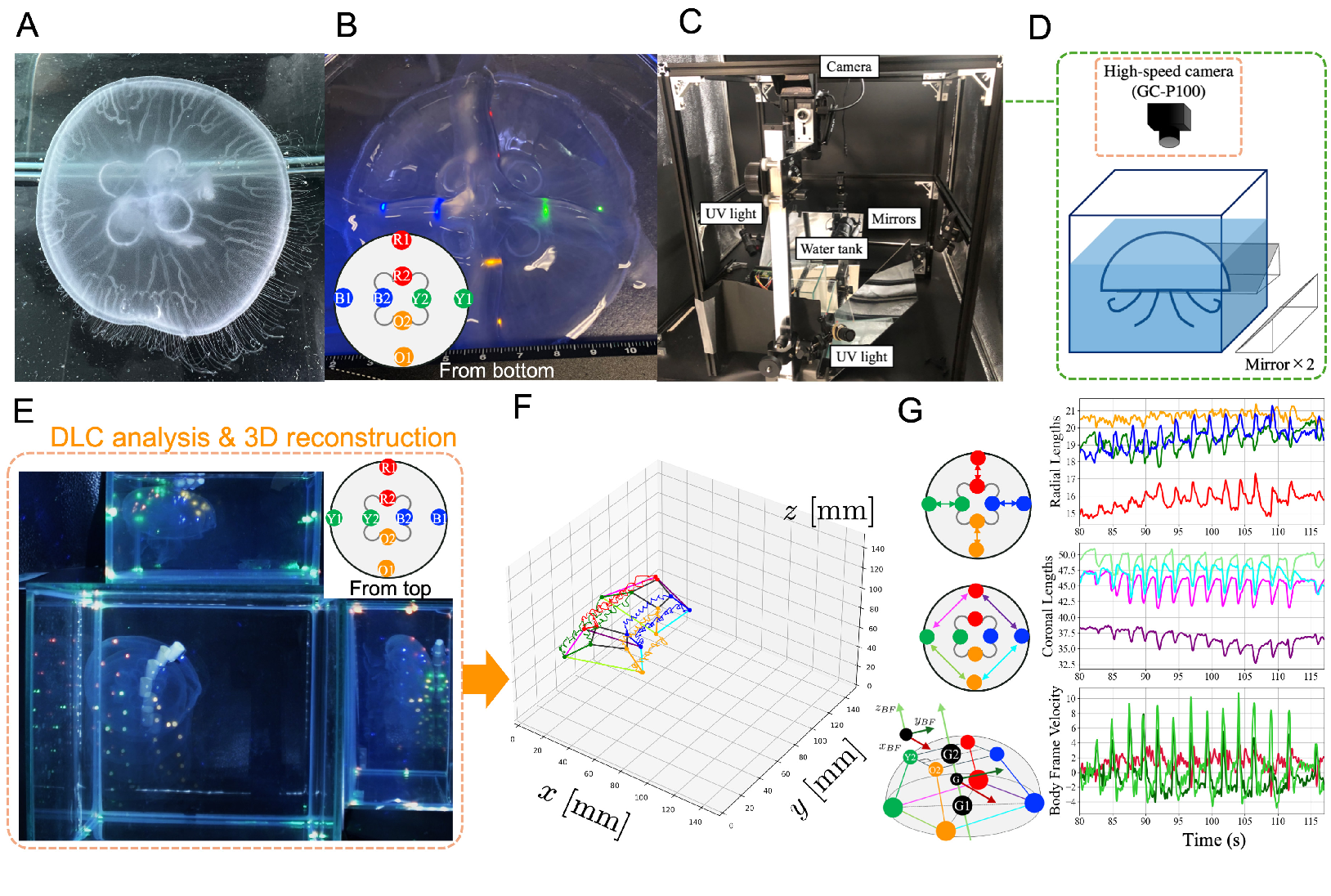}
      \caption{Experimental Animals and Setup: ({\bf A}) {\it Aurelia coerulea} medusae. ({\bf B}) Visible Implant Elastomer Tag under UV light. R*, Y*, O*, and B* in the figure correspond to the colors of the markers, with 1 and 2 indicating the positions of the outer and inner markers, respectively. ({\bf C}) Measurement system with a high-speed camera. ({\bf D}) One camera captured the top view, while two mirrors were positioned on the sides of the tank to enable synchronized, simultaneous recording from three different angles. ({\bf E}) Overlaid snapshots of camera images were taken every 6 seconds. The floating locomotion of jellyfish in a tank was reconstructed and quantified in three-dimensional (3D) space by estimating marker positions using DeepLabCut (DLC, \cite{nath2019using}, Fig. \DLC). ({\bf F}) The 3D motion of the floating jellyfish was reconstructed. ({\bf G}) Time-series data of radial (top) and coronal (middle) lengths on the jellyfish body, along with 3D velocity in the defined BodyFrame (bottom), were calculated and presented (see Method for details).}
      \label{fig:setup}
    %\vspace{-7mm}
\end{figure}

We developed a system to control a biological jellyfish cyborg (Fig. \ref{fig:system}), enabling it to utilize the interaction between its soft body and the water environment to generate efficient, adaptive, and flexible behaviors \cite{megill2002biomechanics,costello2021hydrodynamics,miles2019don,neil2018jet,sinigaglia2020pattern}. 
This system allows for quantitative and statistical data acquisition during stimulus intervention experiments, while closely replicating natural floating behaviors (Fig. \ref{fig:system}A, supplementary video \StimSystem).
Electrodes (Fig. \ref{fig:system}D) were inserted into the underside of the jellyfish's umbrellar body (Fig. \ref{fig:system}C, {\it Aurelia coerulea}, Fig. \ref{fig:setup}A) to provide electrical stimuli to the coronary muscles \cite{Horridge1954The,RomanesXITC,Costello1994Morphology}, which are located the underside of the body.
Pulse Width Modulation (PWM) signals (Fig. \ref{fig:system}B), emulating neural commands to the muscles, were generated using a custom-developed electrostimulation device (Fig. \ref{fig:system}E) and applied to the electrodes to induce muscle contraction and generate pulsatile motion. We systematically investigated the conditions that lead to effective floating locomotion in jellyfish by adjusting the parameters of the PWM signal (see Methods).

To quantify the spatiotemporal response behavior of the jellyfish to electrostimulation and collect data for motion prediction using Reservoir Computing (RC), we developed a custom 3D motion capture system. This system allows us to measure jellyfish behaviors in a three-dimensional space within a water tank (see Fig. \ref{fig:setup}). The key features of our system include: (i) implementation of Visible Implant Elastomer (VIE) tag markers that reflect UV light (Fig. \ref{fig:setup}B) to measure the deformation of the transparent jellyfish body, (ii) recording from three orientations using a top view camera and two mirrors (Fig. \ref{fig:setup}C and D), and (iii) marker position estimation through DeepLabCut \cite{nath2019using} (Fig. \ref{fig:setup}E) and 3D motion reconstruction (Fig. \ref{fig:setup}F). With this system, we are able to quantify the jellyfish's 3D floating trajectory in the tank (Fig. \ref{fig:setup}F), collect time series data on the length between markers on the body, and measure the locomotion speed along the BodyFrame of the jellyfish (Fig. \ref{fig:setup}G) (see Methods for more details). 

\subsection*{Spontaneous Pulsatile Floating Behaviors}
\begin{figure}[t]
  \centering
  \includegraphics[width=.99\linewidth]{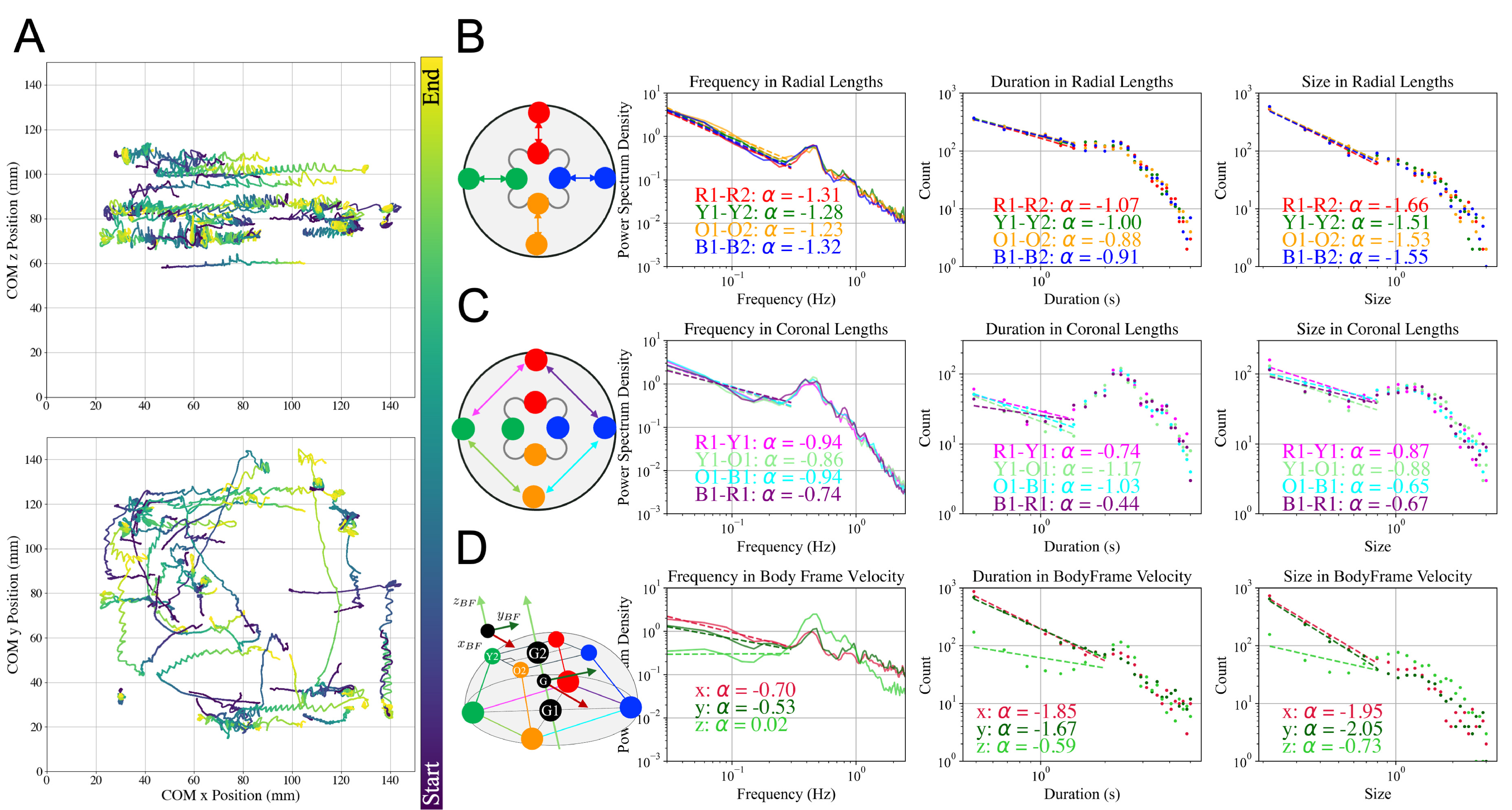}
  \caption{Spontaneous Pulsatile Floating Pattern: ({\bf A}) 3D spontaneous floating trajectories in a water tank ($N=6$ animals, 37 trials). Trajectories in the $x$-$z$ plane (top) and $x$-$y$ plane (bottom). Colors on the maps indicate the start and end of the measurements.  ({\bf B}), ({\bf C}), and ({\bf E}) Distribution of power spectrum density (left column), duration (center column), and size (right column) for radial length (B), coronal length (C), and BodyFrame velocity (D). The vertical and horizontal axes are shown in log scale. The colors repsesent different lengths and velocities: R1-R2 (red), B1-B2 (blue), O1-O2 (orange), Y1-Y2 (green), R1-B1 (purple), B1-O1 (cyan), O1-Y1 (green), Y1-R1 (magenta). $x_{BF}$ velocity (crimson), $y_{BF}$ velocity (dark-green), $z_{BF}$ velocity (lime-green). Each dotted line represents a linear regression line in the corresponding region, and $\alpha$ is the value of the power exponent.} 
  \label{fig:spon_exp}
%\vspace{-7mm}
\end{figure}

Figure \ref{fig:setup}F illustrates the reconstructed 3D spontaneous pulsatile floating behaviors of the jellyfish in the water tank. The color trajectories (red, green, orange, blue) represent the VIE tags' trajectories (R1, R2, O1, O2, Y1, Y2, B1, and B2) in the tank. In Fig. \ref{fig:setup}F, the squares connected by lines demonstrate the deformation of the jellyfish body, connecting R1-Y1-O1-B1 (the outer markers) and R2-Y2-O2-B2 (the inner markers) in the initial and final states, respectively. By comparing with the videos (supplementary video \SpontPattern), we can confirm that the 3D motions can be qualitatively reproduced from the 2D video images.

In Fig. \ref{fig:setup}G, the time evolution of the radial (top) and coronal (middle) lengths calculated from the reconstructed marker positions is displayed. The line colors correspond to those of the jellyfish in the left panels of Fig. \ref{fig:setup}G. We observed active spontaneous oscillations for radial and coronal lengths generated through ring (subumbrellar) muscle expansion and contraction in the jellyfish body. The bottom panel in Fig. \ref{fig:setup}G shows the velocity changes in the $x_{BF}$ (crimson), $y_{BF}$ (dark-green), and $z_{BF}$ (lime-green) directions in the BodyFrame defined for the jellyfish's body (Methods). In the BodyFrame velocity, the increase and decrease of the velocity synchronized with the contraction of the body. The velocity changes in the $y_{BF}$ direction, which mainly corresponds to the direction in which the jellyfish was moving, and in the $z_{BF}$ direction, which corresponds to the direction above the bell of the jellyfish, are also notable.

In Fig. \ref{fig:spon_exp}A, the locomotion trajectories of spontaneous floating patterns in a 3D water tank are displayed ($N=6$ animals, $n=37$ trials). The trajectories are plotted in the $x$-$z$ and $x$-$y$ planes. The measurements were conducted over a relatively long time scale of 35--150 seconds, with the start and end times indicated by the color scale (viridis). The initial position of the jellyfish was random. Notably, the height in the $z$-direction is constrained by the tethered floating system depicted in Fig. \ref{fig:system}C. We observed four types of spontaneous floating patterns: periodic pulsating straight locomotion, periodic pulsating rotational movements, a combination of these movements, and a resting state where the jellyfish remained stationary without pulsating. Some animals were unable to locomote around the walls or corners of the tank due to the limited range and instead continued to pulsate. 
The results indicate that, despite the constraints in the $z$ direction, the jellyfish moved spontaneously in the $x$-$y$ plane with relatively few limitations and exhibited various motion patterns, characterized by intermittent pulsatile oscillations.

To investigate the spontaneous floating pattern of jellyfish, we analyzed the time series data of their body lengths, including radial lengths (B), coronal lengths (C), and BodyFrame velocity (D). In Fig. \ref{fig:spon_exp}B--D, the left column shows the frequency analysis (Power Spectrum Density) of these time series data on a logarithmic scale. 
The plot shows the average including data from all trials for all individuals.
Notable frequency peaks indicate the inherent pulsating motion.
%, which varies among individuals. 
Linear regression lines up to the peaks reveal a power law relationship: (B) $\alpha$ = -1.23 to -1.38; (C) $\alpha$ = -0.74 to -0.94; and (D) $\alpha$ = -0.53 to -0.70, except in the $z_{BF}$-direction, where $\alpha$ = 0.02. 

We extracted pulse-like oscillation waveforms from the time series data to assess the characteristics of self-organized criticality (SOC) \cite{turcotte1999self, Bak1996, Jensen1998, Kagaya2024}. SOC refers to the property of complex systems that naturally evolve into a critical state, characterized by power-law distributions in frequency spectrum, event durations, and sizes. We calculated the duration (burst intervals, in seconds) and size (integral value, area) of the pulses, plotting their distributions in the center and right columns of Fig. \ref{fig:spon_exp}B–D (see Methods for calculation details). The analysis revealed that the radial lengths (B) had durations and sizes with $\alpha$ values ranging from -0.88 to -1.66, the coronal lengths (C) had $\alpha$ values from -0.44 to -1.17, and the BodyFrame velocity (D) had $\alpha$ values from -1.67 to -2.05, except in the $z_{BF}$-direction.

In this study, we demonstrated the potential presence of SOC in the spontaneous floating locomotion of jellyfish, observing power-law relationships in the frequency spectrum and pulse characteristics (duration and size) of their body movements. These findings suggest that jellyfish exhibit spontaneous SOC-driven adaptive behaviors, which could be crucial for understanding the inherent mechanisms of locomotion.

\subsection*{Stimulated Pulsatile Floating Behaviors} 
%
%removed previous figs%
%
\begin{figure}[t]
    \centering
    \includegraphics[width=.99\linewidth]{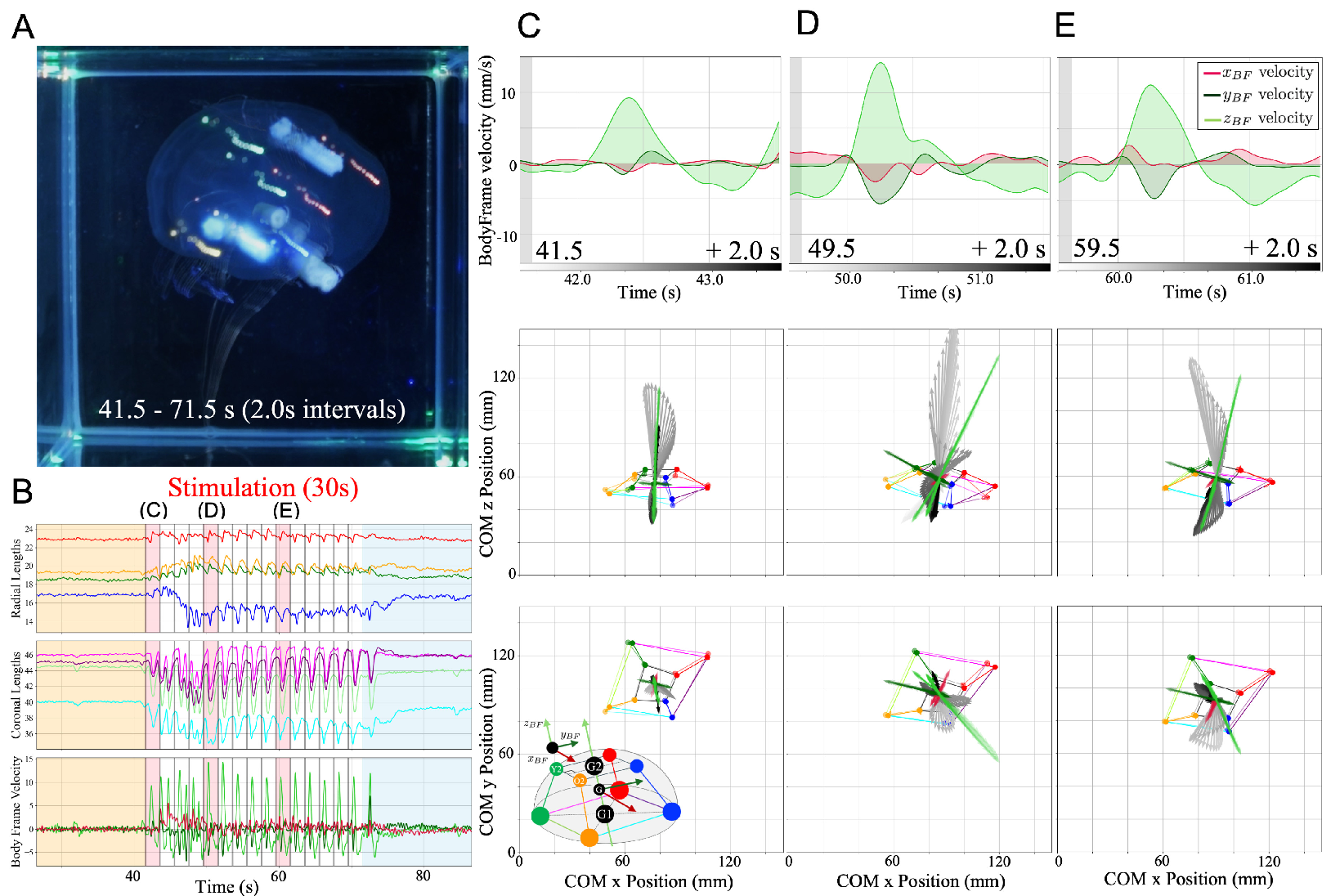}
    \caption{A Representative Example of the Floating Pattern Induced by Electrostimulation $\tau=2.0$ s (supplementary video \StimPattern): ({\bf A}) Locomotion trajectory of a jellyfish (top view) during the 30-second electrostimulation period. Overlaid snapshots are shown at 2.0-second intervals of the stimulation cycle, with higher transparency close to the stimulation start time.  ({\bf B}) Time series data of radial and coronal lengths, as well as BodyFrame velocities, including periods before, during, and after electrostimulation (yellow: before 15 s, white: stimulation 30 s, sky-blue: after 15 s). The gray area within the 30-second stimulus interval represents a duration of 0.1 seconds for a PWM signal input with a 2.0-second period. The pink highlights indicate one cycle (2.0 s) of electrostimulation in (C)-(E) below. ({\bf C}),  ({\bf D}),  ({\bf E}) Top row: Velocity changes in the $x$ (crimson), $y$ (dark-green), and $z$ (lime-green) directions in the BodyFrame during one stimulus cycle (2.0 s). Middle row: Jellyfish's position at the beginning and end of stimulation, as well as the time evolution of the velocity vector during electrostimulation in the $x$-$z$ plane (side view) of the tank. The velocity vectors change in shade of black from the beginning to the end of the 2.0 s stimulation. The velocity vectors in the $x$, $y$, and $z$ axes of the BodyFrame are also shown in each color. Bottom row: Identical data in the middle row, but in the $x$-$y$ plane (top view).}
    \label{fig:stim_exp}
  %\vspace{-7mm}
\end{figure}

Figure \ref{fig:stim_exp} illustrates the floating pattern observed when a 2.0-second periodic electrical stimulus was applied to the electrode between O1-B1. The following parameters of the PWM signal were kept constant in the experiments: frequency ($f$) was set to 50 Hz and duty ratio to 50\%. These parameters were determined to generate effective pulsatile motion through preliminary trial-and-error experiments. To investigate how the pulsatile motion varied by adjusting the stimulation period (Fig. \ref{fig:system}B), we analyzed changes in behavior before and after electrical stimulation. We quantified the floating patterns during three time intervals: 40 seconds before, 30 seconds during, and 40 seconds after the periodic electrical stimulation. As a control condition, we also conducted experiments without stimulation for 110 second intervals (40+30+40) using the floating tethered system and electrodes, referred to as the ``w/o stimulation" condition.

In Fig. \ref{fig:stim_exp}A, the superimposed top view of the camera measuring the locomotion trajectory of the jellyfish during 30 seconds of electrical stimulation with 2.0-second intervals is shown. Figure \ref{fig:stim_exp}B displays the time series data of radial lengths (top), coronal lengths (middle), and BodyFrame velocity (bottom) for $15$ seconds before (yellow) and 15 seconds after stimulation (light blue). We observed a pulsatile pattern during the electrostimulation period, characterized by contraction and increased velocity, mainly in the $y_{BF}$- and $z_{BF}$-directions, synchronized with the input period. Figures \ref{fig:stim_exp}C--E depict the changes in BodyFrame velocity (top), the position of the jellyfish before and after stimulation (center and bottom), and the projection of the BodyFrame velocity vector trajectory onto the $x$-$z$ plane (center) and transverse plane ($x$-$y$) (bottom) during the relevant period of the electrostimulation in Fig. \ref{fig:stim_exp}B (highlighted in pink). The figure illustrates that the velocity change in the $z_{BF}$ direction was mainly induced in the initial stimulus period, while the velocity vector in the $x$-$y$ plane was induced in the latter half of the stimulus period. Specifically, during the 2.0 seconds of the stimulus cycle (especially in E, which settled into a steady pattern), in the first half of the stimulus period just after $0.1$ second of electrostimulation (start to around $1.0$ second, gray arrows), velocity vectors in the direction of travel in the $x$-$y$ plane and the positive direction of $z_{BF}$ were induced. In the second half (from around $1.0$ second to the end, black arrows), on the other hand, the velocity vectors in the opposite direction to the locomotion direction in the transverse plane and in the negative direction of $z_{BF}$ were induced. These changes in the velocity vectors contribute to the increase or decrease in the whole-body velocity of the jellyfish during one stimulus cycle, resulting in muscle stimulus-induced locomotion.

\begin{figure}[t]
  \centering
  \includegraphics[width=.93\linewidth]{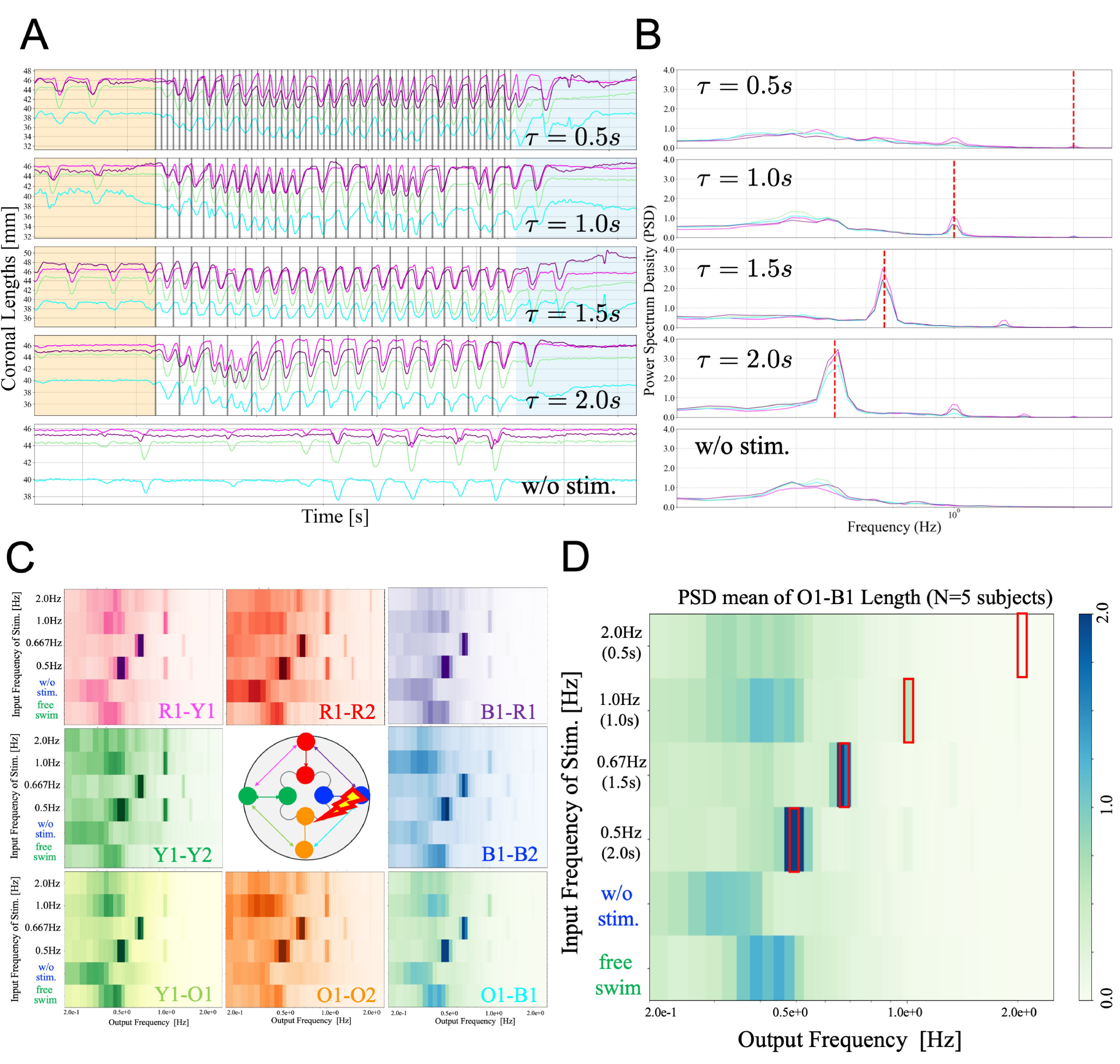}
  \caption{Spatiotemporal Pulsatile Floating Pattern Analysis for Electrostimulation: $N=5$ subjects.  {\bf (A)} Time-series data in four stimulus input conditions: $\tau=0.5$, 1.0, 1.5, 2.0s, and control condition (without stimulation with electrodes and wires). For each condition, we conducted 5 trials, with the order of conditions randomized. To ensure consistency, we fixed the wires between R1 and Y1 and between O1 and B1. We only applied a PWM signal to the electrode between O1 and B1. The gray area within the 30-second stimulus interval represents a duration of 0.1 seconds for the PWM signal input, with each period $\tau$. {\bf (B)} Frequency response analysis of the coronal lengths in four stimulus input (the red dotted lines show input frequencies) and control conditions. {\bf (C)} We analyzed the power spectrum density (PSD distribution) for each spatially distributed radial and coronal length. {\bf (D)} Detailed display of frequency analysis (O1-B1) for four stimulation input conditions, as well as cased without electrostimulation with electrodes and wires (w/o stim column) and spontaneous floating motion without electrodes and wires (free swim column), coresponding to Fig. \ref{fig:spon_exp}. Vertical and horizontal axes indicate input and output frequencies, respectively, and color shading indicates PSD. The red squares represent the input frequencies for the four conditions.}
  \label{fig:stim_exp2}
%\vspace{-7mm}
\end{figure}

Figure \ref{fig:stim_exp2} displays the frequency response of the radial and coronal lengths in the jellyfish body ($N=5$ animals) when exposed to electrical stimulation administered to the ring muscle. The stimulation was delivered via an electrode positioned between O1 and B1. Figure \ref{fig:stim_exp2}A presents time-series data across four stimulus input conditions: $\tau=0.5$, 1.0, 1.5, and 2.0 seconds, along with a control condition (without stimulation using electrodes and wires). For each condition, five trials were conducted, with the order of conditions randomized. To maintain consistency, the two electrodes were secured between R1 and Y1 and between O1 and B1. Figure \ref{fig:stim_exp2}B depicts the Power Spectrum Density (PSD) of the coronal lengths under the four stimulus input conditions (the red dotted lines indicate input frequencies) as well as the control condition. Figure \ref{fig:stim_exp2}D provides a detailed visualization of the frequency response for the lengths O1--B1, similar to panel B. The vertical axis of Fig. \ref{fig:stim_exp2}D represents the frequency of the electrical stimulation input. The resulting frequency responses corresponding to four stimulation input periods are displayed on the horizontal axis using color shading. The ``w/o stimulation" column represents the frequency response in the presence of wires but without electrostimulation (control), while the ``free swim" column indicates the frequency response during spontaneous floating, as shown in Fig. \ref{fig:spon_exp}. The red square in Fig. \ref{fig:stim_exp2}D indicates the input frequency. Finally, Fig. \ref{fig:stim_exp2}C depicts the PSD distribution for each spatially distributed radial and coronal length, with the color map representing the lengths associated with each color. The data in Fig. \ref{fig:stim_exp2}C reveal no discernible trend in the difference between the spatial location of the input stimulus (O1--B1) and the frequency response of each body part.

\begin{figure}[t]
  \centering
  \includegraphics[width=.99\linewidth]{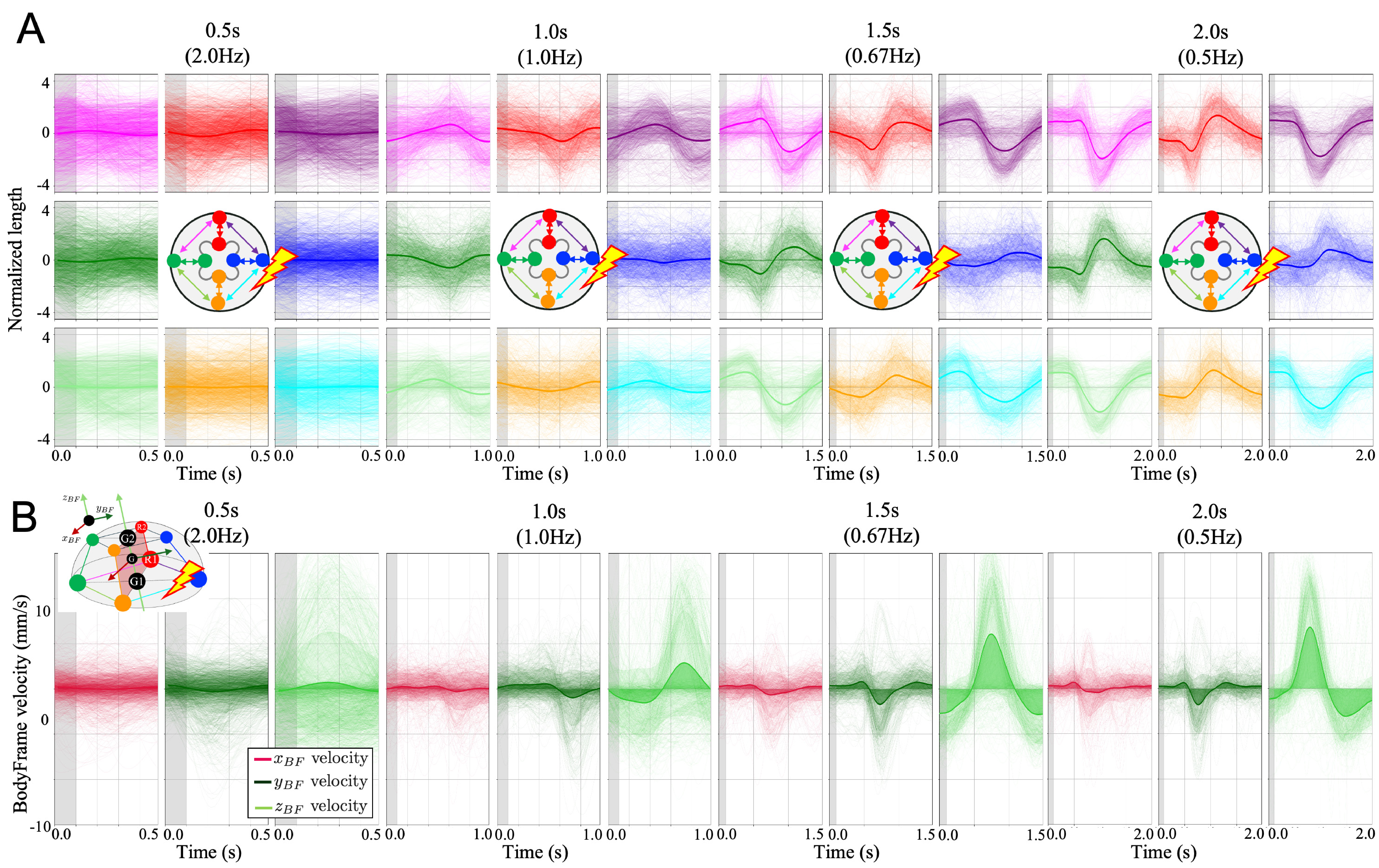}
  \caption{Phase Response for Electrostimulation:  {\bf (A)} Phase response analysis for both radial and coronal length, using normalized length data. The analysis includes date for each condition ($\tau=0.5$ , 1.0, 1.5, and 2.0s). Each time series data set were segmented and overlaid for each input stimulus period, with a total 5 subjects $N=5$ and 5 trials per condition. The mean trajectory is represented by the dark line. {\bf (B)} Phase response analysis of velocity components, $x_{BF}$ (crimson), $y_{BF}$ (dark-green), and $z_{BF}$ (lime-green) velocities on the BodyFrame for each condition (0.5, 1.0, 1.5, and 2.0s). The data extraction and display methods follow a similar approach as in {\bf(A)}. The average velocity represented by a bold line, while the region between the average velocity and velocity 0 is represented by the corresponding transparent color. }
  \label{fig:stim_exp2-2}
%\vspace{-7mm}
\end{figure}
Figure \ref{fig:stim_exp2-2}A illustrates the phase response of body length (radial and coronal lengths) to periodic electrostimulation inputs with periods of $0.5$s, $1.0$s, $1.5$s, and $2.0$s. The plots show the normalized length time variation during the 30s electrostimulation period for each input frequency. The gray shaded regions represent the $0.1$s duration when the PWM signals were injected (Fig. \ref{fig:system}B). The color of each plot corresponds to the length color located on the jellyfish body. The bold lines indicate the average change in length for $N=5$ subjects with 5 trials per condition. Notably, coherent length changes were observed in the $1.5$s and $2.0$s input conditions, coinciding with the frequency peaks depicted in Fig. \ref{fig:stim_exp2}B--D. Contraction and expansion changes in length were clearly observed in these input conditions, suggesting the generation of pulsatile motion synchronized with the input frequency. In contrast to the frequency response, a spatial distribution of the phase response was observed. The phase-time response curve was symmetrical to the line between the jellyfish body center of mass and the electric stimulus point (midpoint of O1 and B1). Thus, the response curves for B1-B2 (blue) and O1-O2 (orange), Y1-O1 (light-green) and B1-R1 (purple), and Y1-Y2 (green) and R1-R2 (red) were identical, respectively. This indicates the generation of a contraction/relaxation progression in the jellyfish body from the stimulation point. Furthermore, the response curves for the $1.5$s and $2.0$s input conditions confirm that the period characterizing contraction/relaxation for each length is approximately $1.6$s.

Figure \ref{fig:stim_exp2-2}B shows the phase response of the BodyFrame velocity to electrostimulation frequency inputs (with $0.5$, $1.0$, $1.5$, and $2.0$s periods). The plotting method is the same as in Fig. \ref{fig:stim_exp2-2} A. The bold lines represent the mean value change for $N=5$ subjects with 5 trials per condition. To clarify positive and negative changes in velocity, the same transparent color is used between the mean velocity and velocity 0. Similar to the findings in Fig. \ref{fig:stim_exp2-2}A, we observed coherent changes in BodyFrame velocity in the $1.5$s and $2.0$s electrostimulation input conditions. Notably, the velocity in the $z_{BF}$ direction was positive in the first half of the stimulus input (around $1.0$s from the start) and negative in the second half (around $1.0s$ to the end). Conversely, the velocity in the $y_{BF}$ direction was negative in the first half and positive in the second half, although not as clearly as $z_{BF}$. In the $1.5$s and $2.0$s conditions with distinct frequency responses, the progressive waves of contraction/relaxation in the jellyfish's soft body induced by muscle electrostimulation resulted in velocity changes in the BodyFrame. This change would contribute to the floating jellyfish's locomotion direction.
\begin{figure}[t]
  \centering
  \includegraphics[width=.99\linewidth]{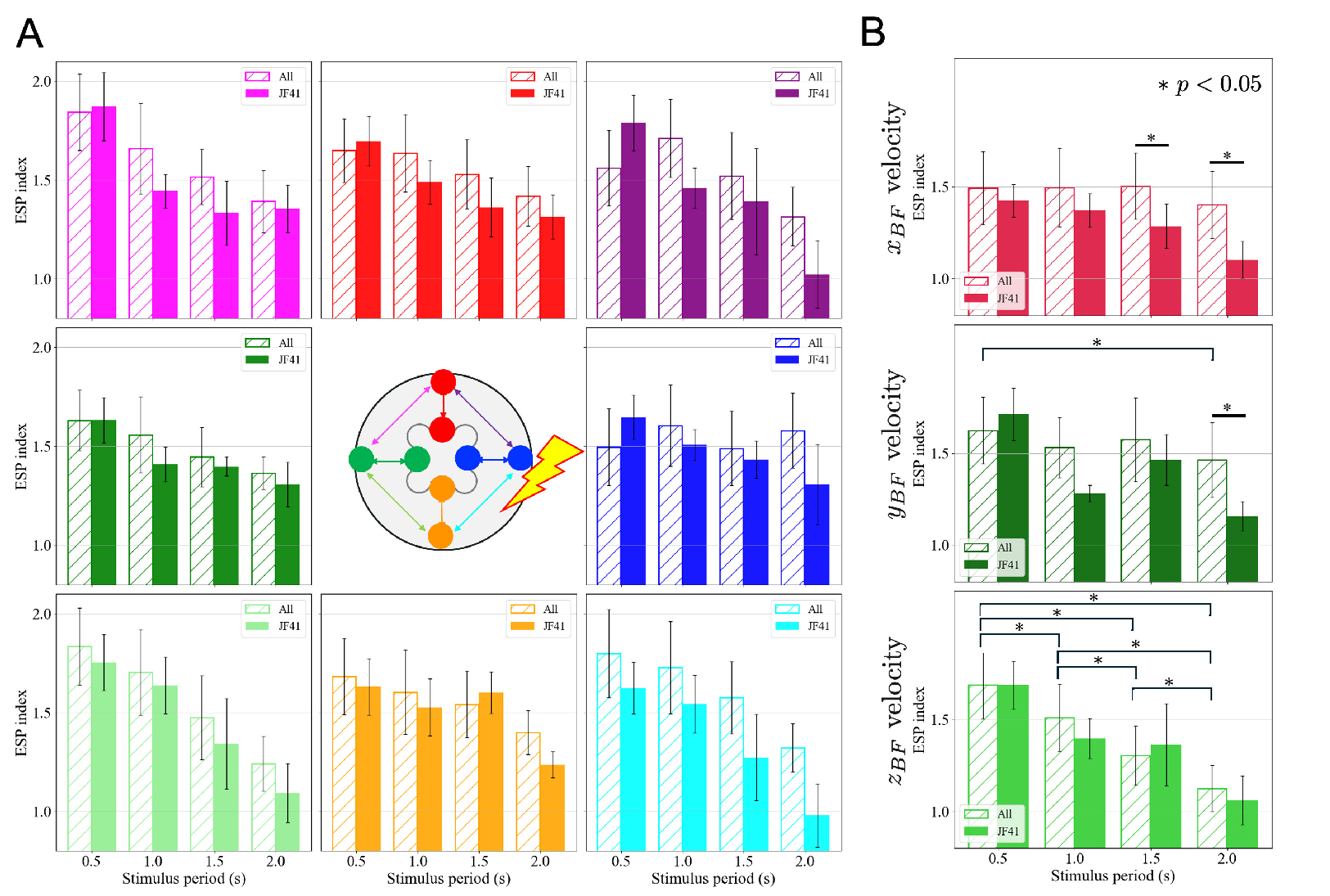}
  \caption{Echo State Property \cite{jaeger2001echo} of Stimulated Floating Patterns: {\bf (A)} Cornal and radial lengths. {\bf (B)} BodyFrame velocities in the $x_{BF}$ (crimson), $y_{BF}$ (dark-green), and $z_{BF}$ (lime-green) directions from the top. We calculated the ESP index for each stimulus period condition (0.5, 1.0, 1.5, and 2.0 s) proposed in \cite{Gallicchio2018Chasing} (see Methods). The white and shaded bars represent the results of the calculations for all 5 jellyfish data (a total of 25 trials), while the color density bars indicate the results of the calculations for one specimen (JF41, 5 trials each). The colors in the graph correspond to the length colors and the BodyFrame speed direction colors. The * in ({\bf B}) indicates that a statistically significant difference was found between the conditions (between stimulus periods in 'All' / between 'All' and 'JF41') using the Tukey HSD test and $t$-test.}
  \label{fig:stim_expESP}
%\vspace{-7mm}
\end{figure}

To quantify the consistency of the phase response to muscle electrostimulation and to determine whether the jellyfish's floating locomotion exhibits computational power as a Reservoir Computing (RC) system, we evaluated the Echo State Property (ESP) \cite{jaeger2001echo} of the input and output data from the stimulus experiments. We calculated the ESP index for each stimulus period condition (0.5, 1.0, 1.5, and 2.0s) using a procedure proposed in \cite{Gallicchio2018Chasing} (see Methods for details). This calculation involved using the Euclidean distance between standardized time series values over a 30-second period of electrostimulation in two trials. A lower ESP index indicates a higher consistency of output in response to input across various initial values, suggesting greater computational power as an RC system. We compared the inter-individual ESP index (using data from all 25 trials for all 5 jellyfish) with the intra-individual ESP index (using data from 5 trials for each jellyfish), as shown in Fig. \ref{fig:stim_expESP} (and Fig. \ESPindex). To examine the differences in the ESP index for BodyFrame velocity as a motion prediction target within the RC framework, we conducted a one-way analysis of variance (ANOVA) to compare the differences between the input periods (0.5, 1.0, 1.5, and 2.0 s). The results indicate statistically significant differences between the input period groups in BodyFrame velocity in the $y$- and $z$-directions ($F(3, 396) = 2.94, p = 0.037$; $F(3, 396) = 52.40, p < 0.001$). Additionally, we conducted Tukey's HSD (Honestly Significant Difference) test for a detailed comparison between each input period. We also performed an independent $t$-test for each period condition to verify the differences between the 'All' (inter-individual data) and 'JF41' (intra-individual data) groups. The * in Fig. \ref{fig:stim_expESP}B indicates that a statistically significant difference ($p<0.05$) was found between the conditions (between stimulus period groups in 'All' / between 'All' and 'JF41' groups) using Tukey's HSD test and $t$-test. 
All $p$-values are provided in Tables \StaticsP~ and \StaticsJ~ in the Supplementary Materials.

The results presented in Fig. \ref{fig:stim_expESP} (and Fig. \ESPindex) highlight three key points: (i) The ESP index is relatively low for the conditions of $\tau = 1.5$ and 2.0 s, indicating potential computational power as a physical jellyfish RC system; (ii) most individuals exhibit lower values in the intra-individual ESP compared to the inter-individual ESP; and (iii) under the condition of $\tau = 2.0$ s, the inter-individual ESP index shows low values only in the $z$ direction, while the intra-individual ESP index, e.g., JF41, displays low values across all $x$, $y$, and $z$ directions. Fig. \ESPindex~ in Supplementary Materials displays the data for all individuals.

\subsection*{Verification of stimulus location and locomotion direction}
\begin{figure}[t]
  \centering
  \includegraphics[width=.99\linewidth]{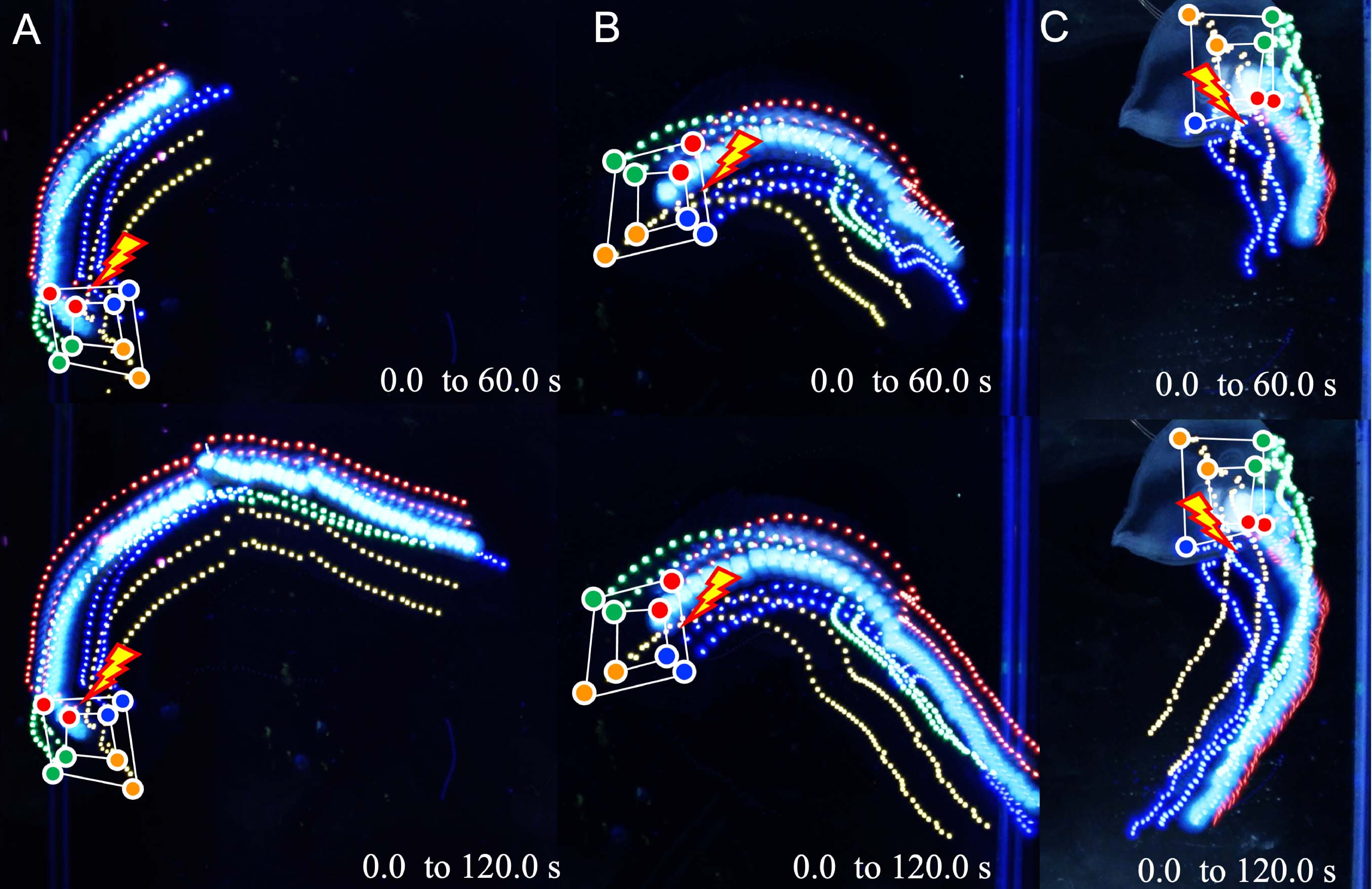}
  \caption{Verification of stimulus location and locomotion direction were conducted using 2.0-s period electrostimulation, which effectively induced pulsatile locomotion (supplementary video \Direction). These experiments lasted for 120 seconds and aimed to determine the relationship between the stimulus input position and the direction of locomotion. The jellyfish locomotion was observed in a water tank with inserted UV reflective markers (VIEs), solely from a top view. For the electrostimulation, only one wire was inserted at the midpoint between R1 and B1. To visualize and plot the trajectories, movies were superimposed at 2.0s intervals. In figures (A), (B), and (C), the trajectories starting from different initial positions and postures (orientations) in the tank are depicted. The upper row displays the trajectory between 0 and 60 s, while the lower row shows the trajectory between 0 and 120 s.}
  \label{fig:dicrect_exp}
%\vspace{-7mm}
\end{figure}
To demonstrate the possibility of controlling the direction of jellyfish locomotion through muscle electrostimulation, we conducted experiments to investigate the relationship between the position of the stimulus input and the direction of locomotion. Using only one electrode, we performed continuous electro-muscle stimulation experiments for 120 seconds. Figure \ref{fig:dicrect_exp} illustrates the locomotion trajectories of jellyfish starting from different initial positions and postures (orientations) in the tank. The upper row shows the trajectory from 0 to 60 seconds, while the lower row shows the trajectory from 0 to 120 seconds. We inserted UV reflective markers (VIEs) into the jellyfish and recorded only the top view. The results in Fig. \ref{fig:dicrect_exp} (supplementary video \Direction) demonstrate that the jellyfish tended to move in the direction of the midpoint between R1 and B1, where the electrostimulation was applied to the muscles, regardless of the initial position and posture of the jellyfish. The curved trajectory during electrostimulation suggests that the wire tension affected the trajectory.

\subsection*{Motion Prediction with Reservoir Computing}
\label{sec:MP}

\begin{figure}[t]
     \centering
     \includegraphics[width=.99\linewidth]{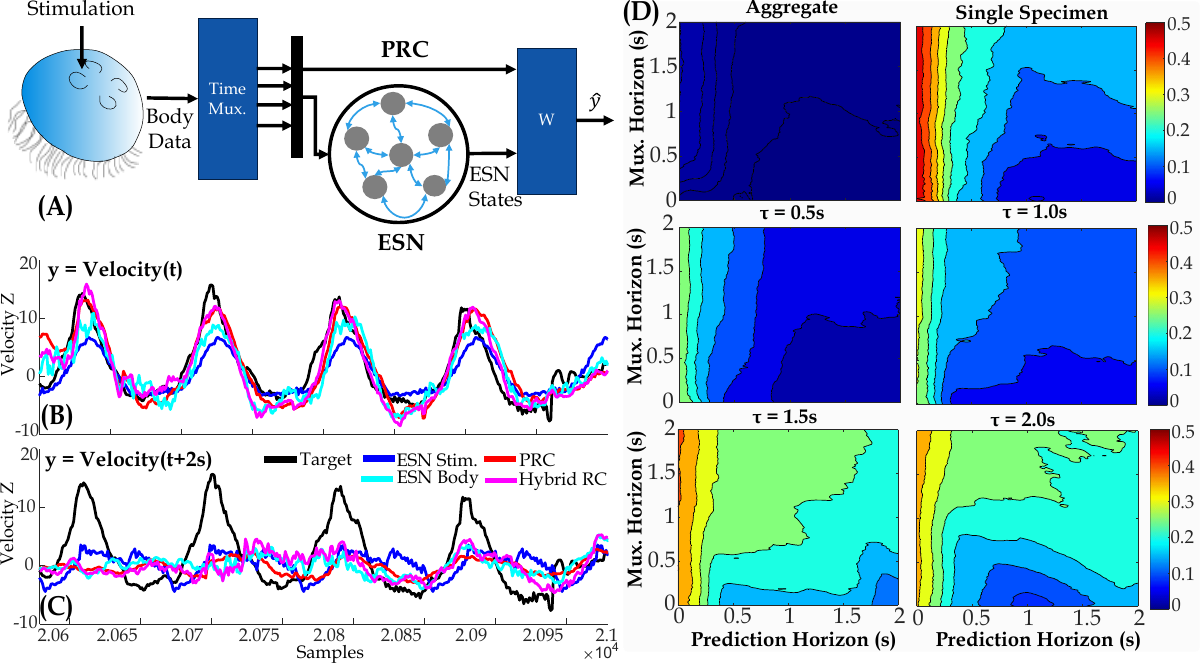}
     \caption{The jellyfish RC structure and data analysis. A) Shows the setup of the ``Hybrid" ESN, with labeled separate elements to illustrate the other RC forms. Examples of the RCs predicting vertical jellyfish velocity are presented for B) the current time and C) $2.0$ s in the future. D) Displays the averaged $R^2$ values for $x$-$y$-$z$ velocities, organized by specimen and stimulus.}
     \label{fig:ESNSetup}
\end{figure}

% \textcolor{red}{$\star \star \star$}
The general form of the Jellyfish Reservoir Computing (RC) system is illustrated in Fig. \ref{fig:ESNSetup}A. Extensive testing was conducted on several different RC forms, synthesizing an Echo State Network (ESN) \cite{jaeger2001echo} and a Physical Reservoir Computing (PRC) with varying inputs. Predictions were made for future Jellyfish motions, including relative position, angular rotations, and BodyFrame velocities (see Methods for RC settings). Sample velocity predictions for various RC structures are presented in Fig. \ref{fig:ESNSetup}B and C for current and future (2.0 s) behaviors. The selected RC structure employs a ``Hybrid'' reservoir computer configuration that combines both ESN and PRC outputs. The jellyfish states input to the RC (i.e., ESN inputs and PRC outputs) derive from an optimal 4-sensor configuration composed of the following lengths: outer radius, inner radius, Y2–O1, and R2–O2 (see Supplementary Material \SupVidESNType ~and \SupMatSensorTesting ~for details on how the sensor and hybrid configuration were chosen).

To evaluate performance using different training data, the Hybrid RC was trained on stimulated pulsatile data aggregated and grouped by animal specimen and stimulus period. Fig. \ref{fig:ESNSetup}D illustrates the averaged $R^2$ values for all velocity targets across aggregated animal data, data from a single representative specimen, and each stimulus period. The horizontal axis represents prediction accuracy for future velocities, while the vertical axis indicates the amount of sensor data history (mux) utilized for the predictions. Notably, the aggregated data (which includes both free and stimulated motions) performs poorly in predicting motions, achieving only marginally better results than a pure regression fit. However, training on body motions during stimulated intervals significantly enhances predictions. The highest average reconstruction accuracy is observed in data organized and trained on input stimuli with $\tau = 1.5$s, followed by $\tau = 2.0$s. These periods also align with the center of mass (COM) and relative marker position profiles that showcase the most consistency across various tests and specimens. These stimuli yield comparable or better accuracy than the best $\tau = 0.5$ s and $\tau = 1.0$s stimuli, even when predicting far into the future with limited mux data. RCs trained on individual specimens exhibit the highest peak performance in velocity. A key difference between data trained on individual specimens and that trained on stimuli is that the former demonstrates higher prediction accuracy across all velocity directions for short horizons, while the latter excels in predicting vertical swimming but struggles with other directions (see supplementary material for supporting data). This trend closely aligns with the motion synchronization patterns observed in the stimulus experiments. 

\begin{figure}[t]
      \centering
      \includegraphics[width=.995\linewidth]{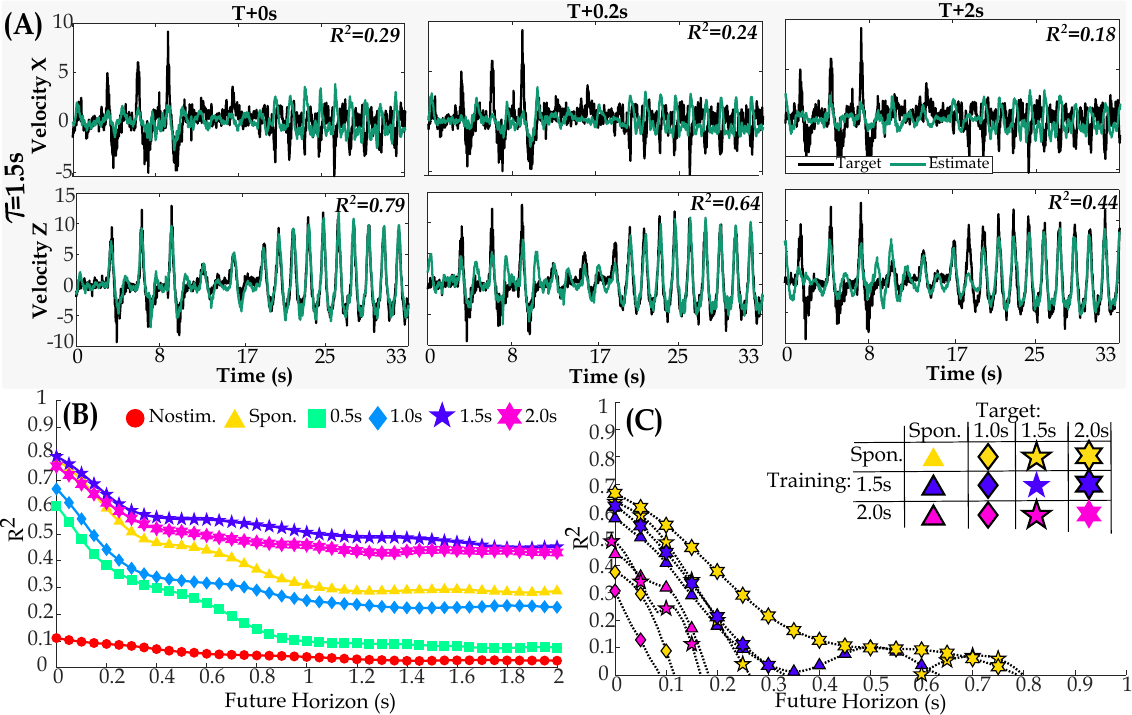}
      \caption{Results of jellyfish motion prediction RC. A) Displays sections of $x$ and $z$ velocity predictions for 0.0 s, 0.2 s, and 2.0 s in the future, using data trained on a $\tau = 1.5$ s stimulus. B) Shows the $R^2$ profile for vertical velocity targets based on data trained on aggregated unstimulated data, collected spontaneous unstimulated pulses, and stimulated pulses. C) Presents a collection of $R^2$ profiles for a confusion task, with targets using training weights from B to estimate other datasets from B.}
      \label{fig:ESNResults}
\end{figure}

To further demonstrate this trend, Fig.\ref{fig:ESNResults}A presents the velocity predictions in both the $x$ and $z$ directions at $\tau = 1.5$s for predictions made over three different future time lengths. Each displayed prediction utilized the maximum input mux length of 2.0s. The predictions in the $x$ direction occasionally follow trends but sometimes show pulses in the opposite direction of the true velocity. In contrast, the $z$ predictions demonstrate a high degree of correspondence with the true trajectories throughout each pulse, even into the future. Instantaneous predictions closely align with actual behaviors in both shape and magnitude. As the RC model predicts further into the future, it appears that the behavioral shape continues to follow trends, but the smaller velocity peaks begin to be overestimated while the larger peaks are underestimated.

To analyze the information encoded in the RC and an aspect of its embodiment, estimations were made by swapping the training and evaluation data sets. The data used included free swimming aggregated data, free swimming spontaneously occurring pulses, and pulsatile responses during each input stimulus period. As a baseline, Fig.\ref{fig:ESNResults}B shows the $R^2$ future prediction profile of the vertical velocity for each trained data set evaluated on its own data. Each of these predictions employed the maximum 2.0s input mux for the RC. While the aggregated data (which includes rest periods) proves very difficult to predict, the spontaneous pulses exhibit a high degree of predictability. The spontaneous pulses have comparable $R^2$ values to both the $\tau = 1.5$s and $\tau = 2.0$s for horizons less than $T+0.3$s and demonstrate better overall predictions than either $\tau = 1.0$s or $\tau = 0.5$s.

The training weights used in Fig.\ref{fig:ESNResults}B are then employed to predict each other's motions. While many of the trajectories were found to predict one another's tasks poorly (particularly those based on the aggregated data), some tasks exhibited surprisingly high performance. For simplicity, Fig.\ref{fig:ESNResults}C presents a selection of these results for $R^2 > 0$ (the complete $R^2$ results and visualizations of the velocity predictions are available in the supplemental material \SupMatConfusion). Data trained on spontaneous pulses was able to predict stimulated data at $\tau = 1.5$s and 2.0s with an $R^2$ as high as 0.68, outperforming predictions for $\tau = 0.5$s pulses. It also demonstrated higher predictability for $\tau = 1.0$s pulses. Furthermore, the $\tau = 1.5$s and 2.0s trained data showed higher predictions for one another as well as for the spontaneous set. The elevated performance of the spontaneous pulses likely stems from their variable time lengths. Stimulated pulses enforce a specific interval, while the timing between spontaneous body contractions is not precisely periodic. Overall, although these showed poorer predictions of long-term behaviors (making them practically less useful than their properly trained counterparts), these results indicate that the best synchronizing stimulated pulses effectively produce dynamics similar to the animals' natural ``embodied'' behaviors.

\section*{Discussion}\label{disc}

%[overall discussion intro]
The goal of this study is to develop a control system for a jellyfish cyborg that effectively harnesses the self-organized adaptive abilities, known as ``embodied intelligence,'' which are naturally embedded in biological animals. The model animal, the jellyfish ({\it Aurelia coerulea} medusa), was selected due to its status as one of the most energetically efficient aquatic animals \cite{miles2019don,neil2018jet}, despite its simple neural structure and the absence of a brain \cite{RomanesXITC,Horridge1954The,Satterlie1985Central,Satterlie2011Do}. This efficiency is achieved through the effective interaction between its soft body deformation and the hydrodynamics of the environment \cite{megill2002biomechanics,costello2021hydrodynamics}, a phenomenon we refer to as ``embodiment \cite{Pfeifer2007SelfOrganizationEA}.'' To investigate this phenomenon, we have developed a floating tethered system that intervenes in jellyfish embodied locomotion by applying electrostimulation to the coronal muscles of their bodies. We have also constructed a custom-built measurement system to quantify and analyze their three-dimensional locomotion in the aquatic environment. Additionally, we applied an extended framework of Reservoir Computing (RC) \cite{Nakajima2020,Nakajima2021} to predict jellyfish locomotion using data accumulated from both spontaneous locomotion and those induced by electrostimulation. This study is the first to successfully incorporate the dynamic motion data generated from the interaction between an animal's body and its environment into an ``embodied'' RC framework aimed at predicting animal locomotion.

%[Spontaneous Patterns]
Using our developed measurement system, we estimated the position of VIE tags injected into the transparent body of jellyfish, allowing us to quantify the spatiotemporal deformation patterns of floating jellyfish that exhibit spontaneous pulsation. Our analysis confirmed the characteristic contraction and relaxation of the jellyfish's soft body shape during spontaneous floating, as well as the associated locomotion velocity along the defined BodyFrame and the time-series data of the macroscopic locomotion trajectory in the 3D aquarium (Fig. \ref{fig:spon_exp}). Frequency analysis of the time-series data revealed two key findings. Firstly, there was a frequency peak (around 0.45 Hz) corresponding to the spontaneous period of jellyfish locomotion. Secondly, a power law was observed in the frequency region up to the peak, except for the $z_{BF}$ velocity. It should be noted here that the body size of the jellyfish used in this study ranged from 49 to 57 mm (see Tab. \ref{tab:jellyfish}), with no large variation, which explains the lack of individual differences in the intrinsic frequency peak of locomotion. The absence of a power law for the $z_{BF}$ velocity, unlike other variables, can be attributed to the weak motion constraint in the $z_{BF}$ direction imposed by the floating tethered system. To further investigate the self-organized criticality (SOC) characteristics \cite{Bak1996, Jensen1998, Kagaya2024} in jellyfish locomotion, we extracted pulse-like patterns from the time-series data and quantified the duration (burst intervals in seconds), as well as its size (integral value or area) (see Fig. \Pulse). As illustrated in Fig. \ref{fig:spon_exp} B-D, the observation of a power law for both the duration and size of the pulses suggests the presence of SOC in the spontaneous floating motion of jellyfish. Previous studies \cite{xu2020low, xu2020field} examined the frequency response of jellyfish muscles to electrostimulation while placed on a dish, but our study is the first to reveal the frequency response during spontaneous floating locomotion. Furthermore, no previous investigations have explored the characteristics of SOC in the spontaneous locomotion of jellyfish, indicating the existence of self-organization mechanisms that are crucial to the animal's adaptive behaviors behind spontaneity.

%[Stimulated Patterns] 
The muscle electrostimulation experiment confirmed that altering the period of the input rectangular pulse via a PWM signal (Fig. \ref{fig:system} B) affected both the pulsatile frequency and BodyFrame velocity (Figs. \ref{fig:stim_exp2} and \ref{fig:stim_exp2-2}). To the best of our knowledge, this study represents the first analysis of frequency and phase response in jellyfish locomotion induced by muscle electrostimulation, as opposed to jellyfish placed on a dish \cite{xu2020low, xu2020field}. The phase response analysis for controllable frequencies (Fig. \ref{fig:stim_exp2-2} A, specifically $\tau=$1.5s and $\tau=$2.0s) provides three notable insights: (i) In the 2.0 seconds period experiment, the characteristic contraction/relaxation time scale of jellyfish muscle induced by electrostimulation is approximately 1.6s. This time constraint accounts for the reduced responsiveness to the $\tau=$0.5s and $\tau=$1.0s stimulus inputs. (ii) The phase response curves for the $\tau=$1.5s and $\tau=$2.0s input periods are symmetrical about the line connecting the center of the jellyfish body and the electrostimulation position (i.e., O1-O2 and B1-B2, O1-Y1 and B1-R1, and Y1-Y2 and R1-R2 are comparable). This symmetry indicates that muscle contraction/relaxation within the jellyfish's soft body propagates spatially from the electrostimulation location toward the body center. (iii) The coherent and consistent phase responses to the $\tau=$1.5s and $\tau=$2.0s inputs, where peaks in frequency response are observed, suggest that the responses exhibit an Echo State Property (ESP) \cite{jaeger2001echo}, a condition applicable for RC in the jellyfish's floating motion. The ESP index analysis \cite{Gallicchio2018Chasing}, shown in Fig. \ref{fig:stim_expESP}, quantitatively demonstrates the potential computational power of the jellyfish RC system. Furthermore, this analysis indicates that intra-individual data show a smaller ESP than inter-individual data, suggesting that intra-individual data may improve the performance of RC learning. Additionally, the ESP analysis of BodyFrame velocity indicates that inter-individual data could predict the BodyFrame $z_{BF}$ direction velocity only, while intra-individual data, e.g., for JF41, could predict velocity in all $x_{BF}$, $y_{BF}$, and $z_{BF}$ directions, suggesting the feasibility of directional control for a jellyfish cyborg with an RC framework. These results correspond to those in Fig. \ref{fig:stim_exp2-2} B, which shows that the phase response of BodyFrame $z_{BF}$ direction velocity has a highly consistent profile across all data for $\tau=1.5$ and 2.0s, while the phase response of the $x_{BF}$ and $y_{BF}$ direction velocities exhibits significant variation depending on the individual. These deviations arise from the influence of electrode weight on the jellyfish’s body shape (as it is challenging to perfectly balance the weight to 0 g in an underwater environment) and the inability to achieve completely uniform and symmetric conditions, such as the positioning of the inserted electrode.

%[ESN]
We have demonstrated that the current state of a jellyfish's body provides information that can be used to estimate its whole-body motion, as well as future movements. Incorporating an Echo State Network (ESN) and/or a number of past body states can significantly enhance these estimations. Estimates based on aggregated data indicate that the jellyfish's body offers little information to ``telegraph'' when a pulse will occur or whether these pulses will involve any maneuvers beyond vertical motion. Spontaneous pulses arise from neural signals, possibly induced by environmental stimuli, but generally appear to be random or exploratory motions that are challenging to predict, aligning with SOC characteristics. However, the motion of the pulses themselves can be predicted fairly well based on visible morphological states. When a stimulus is applied that consistently induces pulsing motions, the swimming behavior of the jellyfish becomes predictable. Furthermore, stimuli that synchronize with the jellyfish's natural spontaneous pulses (such as periods of $\tau = 1.5$s and $\tau = 2.0$s) result in improved predictions and faster swimming. These pulses also serve as a valuable data source for predicting the motions of spontaneous swimming (at least for short horizons of $<0.2$s) and vice versa. Although predictions of vertical motions are accurate, the data collected here make it difficult to estimate rotational motion, and predictions of movements directed in the transverse ($x$-$y$) plane are fair to poor. While tests across specimens did not always result in consistent transverse motion during stimulation, individual specimens often exhibited reliable movements in these directions when stimulated (likely due to variability among animals and electrode placement). Consequently, the estimation of individual animals improved in the transverse plane. This suggests that, in the future, predictions of other directions and rotations may be feasible if a stimulus method can consistently induce these motions across specimens. It remains unclear whether this method matches the accuracy of a high-fidelity fluid-structure model; however, in additional testing, we demonstrated that this model can be executed on a commercially available 3.5 cm$^2$ microcontroller: the SEEED Xiao SAMD21. The results and video of these predictions are available in the supplementary materials \SupMatXiaoTesting ~and \SupVidViableRC. These resources illustrate the relatively low computational load required for a viable real-time control and prediction system. Thus, stimuli and the RC framework pave the way for implementing higher-level control architectures, such as model predictive control, directly on a living jellyfish cyborg while preserving its embodied intelligence.

%[Limitations]
Still, there are several limitations to the experiments in the present study. First, it is not possible to completely eliminate the effects of physical properties such as electrode weight and wire tension needed for electrostimulation in an underwater environment. One possible solution is to use larger individual jellyfish, specifically those larger than 200 mm, where the impact of electrodes and wires is less significant. Another option is to transform the system into a fully mobile cyborg robot, as seen in previous studies \cite{xu2020low,Xu2020EthicsOB}. Second, we have not yet validated the simultaneous or time-delayed electrical stimulation input using multiple electrodes. Initially, our study involved four electrodes attached to the jellyfish body. However, the weight and tension of the electrodes and wires proved to be a significant issue as described above, twice as large compared to using two electrodes. By addressing this problem and implementing electrical stimulation at multiple locations, we can anticipate the possibility of generating complex locomotion trajectories by integrating the findings from this study on one-electrode electrostimulation. Third, our study used a relatively small water tank (150mm$\times$150mm$\times$ 150mm) due to the limitations of the measurement environment. As a result, we faced restrictions on the duration of spontaneous floating locomotion without external environmental input. Furthermore, after stimulation, we observed in some trials that locomotion near the tank wall was almost immobile due to physical constraints. Expanding the measurement system to larger tanks in the future will enable us to gain a better understanding of the characteristics of long-duration and long-distance locomotion. On the other hand, to ensure more systematic data acquisition, we designed a measurement system specialized for the laboratory environment. The primary goal was to accumulate data for a comprehensive understanding of the locomotion principle and its application to an RC framework. The precise data obtained from this measurement system, along with knowledge of frequency response, phase response, velocity characteristics in BodyFrame, and motion prediction using embodied reservoirs, will contribute to the development of a jellyfish cyborg control strategy that maximizes its potential self-organized abilities, such as efficiency and adaptability derived from its embodiment. To successfully develop jellyfish cyborgs that aid in ocean exploration and marine pollution abatement, it is crucial to enable behavioral control according to the desired objective while preserving the jellyfish's self-organized embodied intelligence.

\section*{Methods}\label{method}

\subsection*{Animals}
{\it Aurelia coerulea} medusae were obtained from Kamo Aquarium in Tsuruoka, Yamagata Prefecture, Japan. The specimens were housed in specialized aquariums designed for jellyfish (Kuranetarium, Nettaien, Miyagi Prefecture, Japan) and maintained at a temperature of 21 to 25 $^\circ$C. These aquariums were filled with natural deep seawater, pumped from the ocean at a depth of 800 meters (Najim 800, bluelab, Shizuoka Prefecture, Japan), which helps to keep the jellyfish in a healthy state for an extended period. Data were collected from six animals ($N=6$) during spontaneous floating experiments; of these, five were also used in electrical stimulation experiments ($N=5$) as shown in Tab. \ref{tab:jellyfish}. Additionally, two jellyfish were utilized for locomotion direction experiments in Fig. \ref{fig:dicrect_exp}.

\begin{table}[t]
\caption{\label{tab:jellyfish} List of Jellyfish used in the spontaneous and stimulation experiments. JF** denotes the identification number of the jellyfish. }
\begin{tabular}{c c c c c c}
\toprule
Animals & Date        & size (mm) & weight (g) & Spontaneous &  Stimulated  \\
        &             &           &            &             & 0.5, 1.0, 1.5, 2.0s, w/o \\
\midrule
   JF41 & 2023.10.12. & 57 &  48               & 7~trials    &  5 * 5 trials \\
   JF42 & 2023.10.13. & 49 &  39               & 6~trials    &  5 * 5 trials \\
   JF43 & 2023.10.16. & 55 &  48               & 7~trials    &  5 * 5 trials \\
   JF44 & 2023.10.17. & 56 &  51               & 6~trials    &  5 * 5 trials \\
   JF45 & 2023.10.17. & 50 &  37               & 6~trials    &   ---         \\
   JF46 & 2023.10.17. & 56 &  49               & 5~trials    &  5 * 5 trials \\
\hline \\
{\bf Total} &         &  &                     & {\bf N=6, n=37} &  {\bf N=5, n=5*25=125} 
%\bottomrule
\end{tabular}
\end{table}

\subsection*{Ethics}
Currently, jellyfish research at Tohoku University and the University of Tokyo is not subject to animal care regulations. We strongly support the ethical considerations and responsibilities discussed by \cite{Xu2020EthicsOB} regarding animal experiments, whether they involve invertebrates that do not require specific approval or vertebrates that do. In our jellyfish experiments, we adhered to the principles of harm minimization, precaution, and the 4Rs (reduction, replacement, refinement, and reproduction). Furthermore, we fully agree with the authors that future research on cyborgs, which may explore ethical boundaries not yet fully considered by ethicists and legislators, will require careful examination of both animal welfare and social consequences \cite{Xu2020EthicsOB}.

\subsection*{Experimental Setup}
We constructed a system to measure and quantify the spatiotemporal patterns of three-dimensional (3D) floating movements in a glass water tank measuring 150 mm $\times$ 150 mm $\times$ 150 mm (Fig. \ref{fig:setup}C). The goal was to capture the jellyfish's spontaneous pulsatile floating motion while adhering to measurement constraints. To efficiently analyze the 3D motion, we designed a measurement system that utilized a single camera positioned above the tank and two mirrors affixed to the tank's sides. This setup allowed for synchronized recording from three different angles (Fig. \ref{fig:setup}D). A high-speed camera (GC-P100, JVCKENWOOD Corporation) recorded the jellyfish's motion data (videos), which were stored on an SD memory card in the camera.

The jellyfish's body is mostly transparent, making it difficult to analyze the spatiotemporal patterns of its pulsatile floating motion from videos taken in its natural state. To address this issue, we used biocompatible Visible Implant Elastomer (VIE) Tags \cite{VIE2023}, which are reflective markers visible under UV light. We injected VIE tags in four different colors (red, orange, yellow, and blue) at eight locations (R1, R2, O1, O2, Y1, Y2, B1, and B2: 1 and 2 indicate the outer and inner positions, respectively) on the jellyfish's body, as shown in Fig. \ref{fig:setup}B. These tags served as position markers for subsequent video analysis.

\subsection*{Electrical Stimulation}
We developed a custom-built electrical stimulator for stimulating jellyfish muscles (Fig.\ref{fig:system} E). An extension circuit board was designed for the Raspberry Pi Pico W (Raspberry Pi Foundation), which included 4-channel Pulse Width Modulation (PWM) signal outputs. The parameters of the PWM signals (Fig.\ref{fig:system}B), including frequency $f$ (ranging from 50 to 150 Hz) and duty ratio (0 to 100\%), were controlled using a Raspberry Pi microprocessor. This setup allowed us to investigate the effects of these parameters on motion generation due to muscle stimulation. Specifically, we set the frequency $f$ to 50 Hz, the duty ratio to 50\%, the amplitude to 3.3 V, and the duration to 0.1 s. We systematically analyzed the behaviors produced by muscle contractions induced by bursts of PWM pulses, varying the period [s] of the PWM signal to identify the condition that most effectively and consistently produced pulsating motion.

The two-wire electrodes were constructed from copper wires with a diameter of 50 $\mu$m, platinum rod tips with a diameter of 254.0 $\mu$m (A-M Systems, Sequim, WA, USA), and tiny chip LEDs (TinyLily 10402, TinyCircuits, Akron, OH, USA) used as indicators of electrical stimulation (Fig. \ref{fig:system}D). The weight of the electrodes was offset with foam material to maintain approximately neutral buoyancy. A pair of stimulation electrodes were bilaterally implanted into the subumbrellar tissue of the jellyfish near the bell margin (Fig. \ref{fig:system}A and C, the location at R1-Y1 and O1-B1).

In stimulated floating experiments using electrodes with copper wires, it is mechanically impossible to completely eliminate the influence of wire forces. However, we made efforts to minimize the impact of the wires on jellyfish motion. To achieve this, we developed a floating tethered system that maintains the jellyfish umbrella in a reasonably constant orientation while keeping wire tension as low as possible, as shown in Fig. \ref{fig:system}A and C. In this method, we threaded a fishing line (blue line in Fig. \ref{fig:system}C) through the gelatin (water) part of the jellyfish's body, which does not affect its movement. We then fixed the endpoints of the fishing line to the foam material using UV resin (red ellipsoid in Fig. \ref{fig:system}C). This tethered method allowed the jellyfish to maintain the umbrella orientation within a range that prevented it from flipping upside down while still exhibiting its spontaneous floating behaviors. Fig. \ref{fig:system}A provides a photograph of a jellyfish specimen where the tether system was successfully implemented, producing a natural pulsating floating motion with the wires and electrodes for stimulation.

\subsection*{Data Analysis}

To quantify the pulsatile floating patterns in 3D space based on 2D high-speed camera images (1920 $\times$ 1080 pixels at 60 fps), we employed a pose estimation algorithm called DeepLabCut (DLC) \cite{Mathis2018DeepLabCutMP,Nath2018UsingDF}. First, we marked training data (50-200 frames for each condition, such as spontaneous floating) to train the algorithm. This allowed the algorithm to automatically estimate 12 positions, including the four corners of the tank and 2 $\times$ 4 color markers on the jellyfish (as shown in Fig. \ref{fig:setup}B), from three camera views (top: c, from behind: u, and from the right: r in the mirrors). This resulted in a total of 36 positions (c1, c2, c3, c4, cR1, cR2, cY1, cY2, cO1, cO2, cB1, cB2, u1, u2, u3, u4, uR1, uR2, uY1, uY2, uO1, uO2, uB1, uB2, r1, r2, r3, r4, rR1, rR2, rY1, rY2, rO1, rO2, rB1, and rB2) recorded during each trial of the experiment (Fig. \DLC). The pose estimation error after processing with DeepLabCut was less than 2.0 pixels (1920 $\times$ 1080 video pixels), indicating that the estimation accuracy was sufficient for further analysis.

The marker estimation using DLC provided 2D positions from three different directions. It is important to note that the data obtained from these directions are temporally synchronized because they were captured by a single camera. While methods exist to reconstruct 3D positions from 2D image data, for simplicity in analysis, we approximated the 3D positions using the estimated 2D positions in the $x$, $y$, and $z$ directions. However, since the four corner landmarks of the water tank do not form a perfect rectangle, we applied a homographic transformation to obtain compensated 2D positions, ensuring that the landmarks appeared as an exact rectangle on the water tank.

Using the reconstructed 3D positions of the eight VIE Tags injected into the jellyfish body (R1, R2, Y1, Y2, O1, O2, B1, and B2, Fig. \ref{fig:setup}F), we analyzed the pulsatile floating patterns in the water tank. We calculated all lengths between markers, focusing specifically on the radial lengths (the four lengths between R1 and R2, Y1 and Y2, O1 and O2) and the coronal lengths (the four lengths between R1 and Y1, Y1 and O1, O1 and B1, B1 and R1) to quantify the spatiotemporal floating patterns (Fig. \ref{fig:setup}G left). For analysis and RC training, we extracted relative body motions from the marker position data (Fig. \ref{fig:setup}G bottom). The outer markers (index 1) and inner markers (index 2) were grouped as separate planes (denoted as G1 and G2 respectively) of the jellyfish, with the global center of mass (COM) selected as the center of the inner markers, as these are closer to the thicker center of the bell. We computed the average distances from the COM to the inner and outer marker sets, denoting them as the inner and outer radii. Euler ($z$-$y$-$z$) rotations were used to orient the animal's body data into an identical local BodyFrame (denoted as BF in Fig. \ref{fig:setup}G bottom). The first two rotations aligned the line segment between the centers of the outer and inner planes of markers directly along the vertical $z$ direction, while the final rotation was selected to fix the line segment from Y2 to O2 in the $x$-$z$ plane. For frequency analysis and phase response analysis, we applied a low-pass filter of 3 Hz to all time series data (as no high-frequency movement of 3 Hz or more was observed in the jellyfish motions) and standardized the time series data excluding the BodyFrame velocity.

During the stimulation experiments, we used four LED indicators to synchronize the stimulus input signals with the analyzed video images. A pair of LEDs indicated the left (white LEDs) and right (green LEDs) electrodes, with one LED indicating ``ON" and the other indicating ``OFF" for the simulations (Figs. \ref{fig:system}C and E). By estimating the positions of the LEDs through the DLC analysis, we were able to automatically analyze the synchronization between the timing of the stimulus inputs and the floating motion of the jellyfish.

Fig. \Pulse~depicts the detection method used to analyze pulse duration and size in relation to self-organized criticality (SOC), as shown in Fig. \ref{fig:spon_exp}. To illustrate this, we consider a case involving periodic B1-R1 expansion and contraction motion with a normalized length. In this scenario, we establish a threshold and define the duration, or burst interval, which is calculated as the time from when the pulse first exceeds the threshold to when it next exceeds it again. Additionally, we calculate the integral (area) of the values above the threshold to represent the size. By analyzing these duration and size metrics, we can evaluate the likelihood of SOC characteristics.
\subsection*{Echo State Property Index Calculation}
We calculated the Echo State Property (ESP) for each stimulus period condition (0.5, 1.0, 1.5, and 2.0s) using the ESP index proposed in \cite{Gallicchio2018Chasing}. This calculation involved using the Euclidean distance between standardized time series values over a 30-second period of electrostimulation in two trials, as described by the following equation:
$$ \delta_{i} (t-T) = || \hat{F} (\bm{x}_{0}, [u(1),\ldots,u(t)]) - \hat{F} (\bm{z}_{0,i}, [u(1),\ldots,u(t)]) ||, $$
where $\hat{F}(\bm{x}_{0}, [u(1),\ldots,u(t)])$ represents the state (radial and colonal lengths/BodyFrame velocities) after the $t$-second muscle electrostimulation input $u(t)$ from the reference initial state $\bm{x}_0$. The initial state of the second term on the right side of the equation was set to $\bm{z}_{0,i}$, and the average value was calculated as follows when $t$ varied from $T+1$ to $L=30$ (the stimulus input time of 30 seconds):
%$$ \var Delta_{i} = \frac{1}{L-T} \Sigma_{j=1}^{L-T}\delta_{i}(j)$$
$$\varDelta_{i} = \dfrac{1}{L-T} \sum_{j=1}^{L-T} \delta_{i}(j), $$
where $T$ is a parameter that determines the transient state period (from $t=0$ to $T$) during which the state deviation (Euclidean distance) is not considered in the calculation. Additionally, we used time-series data for $P=25-1$ for all individuals or $P=5-1$ for intra-individual variations with different initial values of $z_{0,i}$ and calculated the average of these values as the ESP index to evaluate the ESP:
$$ \mathrm{EPS~index} = \dfrac{1}{P} \sum_{i=1}^{P} \varDelta_{i} .$$
We also varied the reference initial value $\bm{x}_{0}$ according to the number of experimental trials (25 or 5) and plotted the averaged value as the ESP index in Fig. \ref{fig:stim_expESP}, excluding duplicate calculations.

\subsection*{Reservoir Computing}
 Possible inputs to the RC consisted of any of the 28 marker-to-marker length the inner/outer body radii, or the stimulation signal to the Jellyfish. These states could either be directly fed to the output state (in the Physical Reservoir computing (PRC) form) and/or used as the input to an Echo State Network (ESN) \cite{jaeger2001echo}. The basic equations for this are expressed below. 
 $$X(T+1)= f(AU(T)+BX(T))$$
 $X(T)$ is the current state of the ESN, $U(T)$ is the input to the ESN, and the activation function is $f(x)=tanh(x)$. $A$ and $B$ are the random input and internal weight matrices respectively. Targets for the RC were separated for pulsatile data sets (which only included data while the animal is pulsing its body) and aggregate data sets (all data including when the animal was at rest). Pulsatile sets included the bodies 3D local frame velocities and dead-reckoned position states (both Cartesian positions and Euler rotations) that were reset to zero at the onset of each pulse to remove the net motion and initial position variance that cannot be estimated from an RC. Aggregate data targets were limited to just the local velocities. Pulses of the jellyfish were extracted by using both the periodic input signal (for stimulated pulses) and based on sharp continuous changes in velocity (for spontaneous pulses). Velocity are computed from position and time data using an explicit-Euler approximation and smoothed with a 5 sample moving average filter. The weight matrix for each of these targets are trained by linear regression.  

As there are more than $155$ million unique sensor configurations of just from the internal body lengths, a best sensor group was determined using the PRC. All possible unique combinations of between 1 and 5 sensors (174,436) were tested in the PRC tasks for emulating the pulsatile targets, aggregate targets, and the input signal, with the best performing sensor configuration recorded for each task. After which the frequency of each sensor's appearance in a best configuration was computed and used to determine which sensors are the best overall (for greater detail see supplementary material \SupMatSensorTesting). The 4 most frequently appearing lengths were chosen to be the best sensor as it was assumed that the animals would have limited capacity to support a large number of on board sensors and this also appeared to allow a fair comparison to other naive configurations during the more rigorous RC testing phase. The spectral radius of the ESN was also initially swept from [0.05,1] at 0.05 intervals and finally chosen to be fixed to 0.35 as it was found to have the best accuracy at all of the target reconstruction tasks with each tested sensor configuration. 

Full testing of the RC focused on predicting current and future behaviors of the animal with the regression estimating the target data from [0,2] seconds in the future on a shifting horizon. To improve these estimates, temporal multiplexing with a shifting horizon of the past input signals was used to explicitly encode additional memory into the RC, with the hope that any periodicity present in these animal may improve the estimation of longer horizons (see supplementary video \SupVidESNType). This multiplexing (mux) appeared in the PRC as additional state trajectories and in the ESN as a leaky-integrator\cite{jaeger2007optimization,Nakajima2018} on the input. As expressed below,
 $$U(T) = [S_1(T,T-1,...,T-\tau_m),S_2(T,T-1,...,T-\tau_),...,S_n(T,T-1,...,T-\tau_m)]$$
the input $U(T)$ is composed of sensor (or stimulation signal) data from sensors $S_1$ to $S_n$ and includes data from the current time $T$ to multiplexed horizon $T-\tau_m$. The objective for these is $Y(T)$ and the approximate is $\hat{Y}$. The approximate is computed with: $\hat{Y} = W[X(t),1]$ for the ESN, $\hat{Y} = W[U(T),1]$ for the PRC, and with $\hat{Y} = W[X(t),U(T),1]$ for the "hybrid" RC where $W$ is the trained weight matrix and 1 is a bias term. Note that, during multiplexing the magnitude of input sensor series is scaled such that the maximum possible sum of input to the ESN has a magnitude of 1 (ie $|\Sigma U(T)| \leq 1$). Furthermore, each new input signal to a node of the ESN caused by the mux has a unique new weights with all other input weights remaining constant (ie $A$ is constant across length of $\tau_m$). The washout phase of the RC was $10^4$ and $10^3$ samples for training with data from aggregated and pulsatile data sets respectively. Due to variability between sensor placement in and size of each specimen there was no separation between the training and evaluation phases. 

\backmatter

\section*{Data availability}
All the data and methods needed to evaluate the conclusions of this work are presented in the main text and Supplementary Information. Additional data can be requested from the corresponding author.

\section*{Code availability}
Codes that are used in this paper are are available from the corresponding author on reasonable request.

%\begin{appendices}
%
%\section{Section title of first appendix}\label{secA1}
%
%An appendix contains supplementary information that is not an essential part of the text itself but which may be helpful in providing a more comprehensive understanding of the research problem or it is information that is too cumbersome to be included in the body of the paper.
%
%\end{appendices}

\bibliography{natcom}% common bib file

\section*{Acknowledgements}
This work was supported by a JSPS KAKENHI (JP23K18472) and by JKA and its promotion funds from KEIRIN RACE.

\section*{Author contributions}

D.O., M.A., and K.N. conceived the basic concept of the jellyfish cyborg. S.I. and K.O. bred and provided the experimental jellyfish. D.O. developed the 3D motion capture and cyborg system. D.O., S.I., and K.O. conducted the animal experiments and collected the experimental data. D.O. and M.A. performed the data analysis. M.A. and K.N. developed the hybrid reservoir computing framework and software implementation for the motion prediction experiment. D.O. and K.N. supervised the project. D.O. and M.A. wrote the initial draft of the manuscript. All authors discussed the results and contributed to writing the manuscript.

\section*{Competing interests}
The Authors declare no competing interests.

\section*{Additional information}
\bmhead{Supplementary information}
%The online version contains supplementary material available at ...

%If your article has accompanying supplementary file/s please state so here. 
%
%Authors reporting data from electrophoretic gels and blots should supply the full unprocessed scans for key as part of their Supplementary information. This may be requested by the editorial team/s if it is missing.
%
%Please refer to Journal-level guidance for any specific requirements.

\bmhead{Correspondence}
Requests for materials should be addressed to Dai Owaki, Max Austin, or Kohei Nakajima.

\end{document}